\pdfoutput=1

\documentclass[11pt]{article}

\usepackage[]{acl}

\usepackage{times}
\usepackage{latexsym}

\usepackage[T1]{fontenc}

\usepackage[utf8]{inputenc}

\usepackage{microtype}

\usepackage{inconsolata}

\usepackage{graphicx}

\usepackage{nicefrac}       
\usepackage{url}

\usepackage{amsthm, mathtools, amsfonts, amssymb, color, relsize, enumerate, enumitem, amsmath, nicematrix, physics}

\usepackage{stackengine}    

\usepackage{subfloat}
\usepackage{algorithm}
\usepackage{algpseudocode}

\renewcommand{\Comment}[1]{\textcolor{navy}{$\triangleright$ #1}}

\usepackage{geometry}
\usepackage{setspace}

\newtheoremstyle{bfnote}%
{}{}                   
{\normalfont}         
{}                    
{\bfseries}          
{.}                  
{ }                  
{\thmname{#1}\thmnumber{ #2}\thmnote{ (#3)}}
\theoremstyle{bfnote}

\newtheorem{theorem}{Theorem}[section]

\newtheorem{definition}{Definition}[section]
\newtheorem{corollary}{Corollary}[section]
\newtheorem{remark}{Remark}[section]

\newtheorem{observation}{Observation}[section]

\usepackage{listings}
\usepackage{appendix}
\usepackage{rotating}

\usepackage[]{caption}
\usepackage[]{subfig}

\usepackage{colortbl, tabularx}

\usepackage[c]{esvect}

\usepackage{bm}

\usepackage{xcolor}

\usepackage{pdfpages}

\usepackage{multirow}
\usepackage{booktabs}

\usepackage[nameinlink, capitalise, noabbrev]{cleveref}
\crefname{section}{Sec.}{Secs.}
\Crefname{equation}{eq.}{eqs.}
\Crefname{figure}{Fig.}{Figs.}
\Crefname{tabular}{Tab.}{Tabs.}
\Crefname{table}{Tab.}{Tabs.}
\crefname{listing}{Algorithm.}{Algorithms.}
\crefname{appendix}{App.}{Apps.}
\creflabelformat{equation}{#2\textup{#1}#3}  

\definecolor{navy}{HTML}{0a1172}
\definecolor{violet}{HTML}{710193}
\definecolor{palegrey}{HTML}{f8f8f9}
\definecolor{deeppink}{HTML}{cb0162}
\definecolor{ferrari}{HTML}{ff2800}
\definecolor{usflag}{HTML}{bf0a30}
\definecolor{strings}{rgb}{.624,.251,.259}
\definecolor{keywords}{rgb}{.224,.451,.686}
\definecolor{comment}{rgb}{.322,.451,.322}

\usepackage{authblk}

\title{Generalized Attention Flow: Feature Attribution for Transformer Models via Maximum Flow}

\author[1,2]{Behrooz Azarkhalili}
\author[2]{Maxwell Libbrecht}
\affil[1]{Computing Science Department, Simon Fraser University}
\affil[2]{Life Language Processing Lab, University of California, Berkeley}

\makeatletter
\renewcommand\AB@affilsepx{, \protect\Affilfont}
\makeatother

\affil[ ]{\texttt{\{bazarkha, maxwell$\_$libbrecht\}@sfu.ca}}

\begin{document}
\maketitle

\begin{abstract} \label{abstract}
	This paper introduces Generalized Attention Flow (GAF), a novel feature attribution method for Transformer-based models to address the limitations of current approaches. By extending Attention Flow and replacing attention weights with the generalized Information Tensor, GAF integrates attention weights, their gradients, the maximum flow problem, and the barrier method to enhance the performance of feature attributions. The proposed method exhibits key theoretical properties and mitigates the shortcomings of prior techniques that rely solely on simple aggregation of attention weights. Our comprehensive benchmarking on sequence classification tasks demonstrates that a specific variant of GAF consistently outperforms state-of-the-art feature attribution methods in most evaluation settings, providing a more reliable interpretation of Transformer model outputs.
	
\end{abstract}

\section{Introduction} \label{sec:1}
Feature attribution methods are essential to develop interpretable machine and deep learning models. These methods assign a score to each input feature, quantifying its contribution to the model's output and thereby enhancing the understanding of model predictions.

\smallskip

The rise of Transformer models with self-attention mechanism  has driven the need for feature attribution methods for interpreting these models \citep{vaswani2017, bahdanau2016, devlin2019, sanh2020, kobayashi2021}. Initially, attention weights were considered potential feature attributions, but recent studies have questioned their effectiveness in explaining deep neural networks \citep{abnar2020b, jain2019, serrano2019a}. Consequently, various post hoc techniques have been developed to compute feature attributions in Transformer models.

\smallskip

Recent advancements in XAI have introduced numerous gradient-based methods, including Grads and AttGrads \citep{barkan2021}, which leverage saliency to interpret Transformer outputs. \citet{qiang2022} proposed AttCAT, integrating features, their gradients, and attention weights to quantify input influence on model outputs. Yet, many of these techniques still focus primarily on the gradients of attention weights and inherit the limitations of earlier attention-based approaches.

\smallskip

Layer-wise Relevance Propagation (LRP) \citep{bach2015a, voita2019a} transfers relevance scores from output to input. \citet{chefer2021a, chefer2021d} proposed a comprehensive methodology enabling information propagation through all Transformer components. Yet, this approach relies on specific LRP rules, limiting its applicability across various Transformer architectures.

\smallskip

Many existing methods to evaluate feature attributions in Transformers fail to capture pairwise interactions among features. This limitation arises from the independent computation of importance scores, which neglects feature interactions. For example, when calculating gradients of attention weights, they propagate directly from the output to the individual input feature, ignoring interactions. Additionally, many methods applied to compute feature attributions in Transformers violate pivotal axioms such as symmetry, sensitivity, efficiency, and linearity \citep{shapley1952, sundararajan2017, sundararajan2020} (\cref{sec:3:subsec:5}).

\smallskip

\citet{abnar2020b} recently introduced Attention Flow to overcome these limitations in XAI methods. Attention Flow considers attention weights as capacities in a maximum flow problem and compute feature attributions using its solution. This approach naturally captures the influence of attention mechanisms, as the paths of high attention through a network correspond to the flow of information from features to outputs. Applicable to any encoder-only Transformer, Attention Flow has demonstrated strong potential to improve model interpretability \citep{abnar2020b, modarressi2023, kobayashi2020c, kobayashi2021}.

\smallskip

Subsequently, \citet{ethayarajh2021b} attempted to bridge attention flows and XAI by leveraging Shapley values \citep{shapley1952, shapley2016}. While their goal was to demonstrate that Attention Flows can be interpreted as Shapley values under specific conditions, they overlooked the issue of non-uniqueness in such flows (\cref{sec:3:subsec:3}).

\smallskip

\textbf{Our contributions.} 
In this work, we propose Generalized Attention Flow (GAF), a method that not only satisfies crucial theoretical properties but also demonstrates improved empirical performance. The primary contributions of our work are:

\smallskip

1. We proposed Generalized Attention Flow, which generates feature attributions by utilizing the log barrier method to solve a regularized maximum flow problem within a capacity network derived from functions applied to attention weights. Rather than defining capacities solely based on attention weights, we will introduce alternatives using the gradients of these weights (GF) or the product of attention weights and their gradients (AGF).

\smallskip

2. We address the non-uniqueness issue in Attention Flow, which previously undermined some of its proposed theoretical properties \citep{ethayarajh2021b}, and demonstrate that non-unique solutions are frequent in practice. To resolve this, we introduce barrier regularization, proving that feature attributions obtained from the regularized maximum flow problem are Shapley values and satisfy the axioms of efficiency, symmetry, nullity, and linearity \citep{shapley1952, shapley2016,young1985, chen2023e}.

\smallskip

3. We conduct extensive benchmarking of the proposed attribution methods based on Generalized Attention Flow, comparing them against various state-of-the-art attribution techniques. Our results show that a specific variant of the proposed method outperforms previous methods for classification tasks across most evaluation scenarios, as measured by AOPC \citep{barkan2021, nguyen2018a, chen2020a}, LOdds \citep{chen2020a, shrikumar2018}, and classification metrics.

\smallskip

4. We have developed an open-source Python package to compute feature attributions leveraging Generalized Attention Flow. This package is highly flexible, and can compute the feature attributions of any encoder-only Transformer model available in the Hugging Face Transformers package \citep{wolf2020}. Moreover, our methods are easily adaptable for a variety of NLP tasks.

\smallskip

\section{Preliminaries} \label{sec:2}
\subsection{Multi-Head Attention Mechanism}\label{sec:2:subsec1}

Given the input sequence $\bm{X} \! \in \!  \mathbb{R}^{t \times d}$, where $d$ is the dimensionality of the model's input vectors and $t$ is the number of tokens, the multi-head self-attention mechanism computes attention weights for each element in the sequence employing the following steps:
\begin{itemize}
	\item \textbf{Linear Transformation}:
	\begin{equation}
		\bm{Q}_i = \bm{XW}^Q_{i}, \hspace{2pt} \bm{K}_i = \bm{XW}^K_{i},  \hspace{2pt} \bm{V}_i = \bm{XW}^V_{i}
		\label{eq:1}
	\end{equation}
	
	Here $\bm{Q}_i, \bm{K}_i \! \in \!  \mathbb{R}^{t \times d_k}$ and $ \bm{V}_i \! \in \!  \mathbb{R}^{t \times d_v}$, where $d_k$ and $d_v$ represent the dimensionality of the key vector and value vector respectively, and $i$ represents the index of the attention head.
	
	\item \textbf{Scaled Dot-Product Attention}:
	\begin{equation}
		\bm{A}^{\! \ast}_{i}(\bm{Q}_i, \bm{K}_i, \bm{V}_i) = \widetilde{\bm{A}}_i \bm{V}_i
		\label{eq:2}
	\end{equation}
	where the  matrix  of attention weights $\widetilde{\bm{A}}_i \hspace{-4pt} \in \hspace{-4pt}  \mathbb{R}^{t \times t}$ is  defined as:
	\begin{equation}
		\widetilde{\bm{A}}_i = \text{softmax}\left(\frac{\bm{Q}_i\bm{K}_i^T}{\sqrt{d_k}}\right)
		\label{eq:3}
	\end{equation}
	
	\item \textbf{Concatenation and Linear Projection}:
	\begin{equation}
		\textbf{MultiHead}(\bm{X}) = \text{Concat}(\bm{A}^{\! \ast}_1, \ldots, \bm{A}^{\! \ast}_h)\bm{W}^O
		\label{eq:4}
	\end{equation}
	where the matrix $\textbf{MultiHead}(\bm{X}) \hspace{-2pt} \in \hspace{-2pt}  \mathbb{R}^{t \times d}$ and the matrix $\bm{W}^O \!  \in \! \mathbb{R}^{h \cdot d_v \times d}$. 
	
\end{itemize}

For a Transformer with $l$ attention layers, the attention weights at each layer can be defined as multi-head attention weights:
\begin{equation}
	\bm{\widehat{A}} = \text{Concat}(\widetilde{\bm{A}}_1, \widetilde{\bm{A}}_2, \ldots, \widetilde{\bm{A}}_h) \! \in \!  \mathbb{R}^{h \times t \times t}
	\label{eq:5}
\end{equation}
Extending this to a Transformer architecture itself, the Transformer attention weights $\bm{A}$ can be defined as:
\begin{equation}
	\bm{A}=\text{Concat}(\bm{\widehat{A}}_1, \bm{\widehat{A}}_2, \ldots, \bm{\widehat{A}}_l) \! \in \!  \mathbb{R}^{l \times h \times 	t \times t}
	\label{eq:6}
\end{equation}

where $\widehat{A}_j \! \in \!  \mathbb{R}^{h \times t \times t}$ is the multi-head attention weight for the $j$-th attention layer.

\subsection{Minimum-Cost Circulation \& Maximum Flow Problem} \label{sec:2:subsec2}

\begin{definition}[Minimum Cost Circulation]
	Given a network $G=(V, E, \bm{u}, \bm{l}, \bm{c})$ with $ |V|=n $ vertices and $ |E|=m $ edges, where $c_{i j}$ is the cost, $l_{i, j}$ and $u_{i, j}$ are respectively the lower and upper capacities (or demands) for the edge $(i,j)\! \in \! E$, a circulation is a function $f: E \rightarrow \mathbb{R}^{\geq 0}$ s.t.
	\begin{equation}
		\arraycolsep=3pt
		\def\arraystretch{1.5}
		\begin{array}{ll}
			l_{i j} \leq f_{i j} \leq u_{i j}, & \forall(i, j) \in E \\
			\mathlarger{\mathlarger{\sum\limits_{j:(i, j) \in E}}} f_{i j} - \hspace{-3pt} \mathlarger{\mathlarger{\sum\limits_{j:(j, i) \in E}}} f_{j i} = 0, & \forall i \in V .
		\end{array}
		\label{eq:7}
	\end{equation}
	The min-cost circulation problem is to compute the circulation $f$ minimizing the cost function $\mathlarger{\sum_{(i, j) \in E}} c_{i j} f_{i j}$. 
	
\end{definition}

The minimum-cost circulation problem can be algebraically written as the following primal-dual linear programming (LP) problem \citep{vandenbrand2021a, chen2023f}:
\begin{equation}
	\begin{split}
		\text{(Primal)} & \underset{\substack{\bm{B}^{\top} \hspace{-2pt} \bm{f}=\bm{0} \\ l_e \leq f_e \leq u_e \forall e \in E}}{\arg \min } \hspace{-4pt} \bm{c}^{\top} \hspace{-2pt} \bm{f} \quad  \text{i.e.} \quad \underset{\substack{\bm{B}^{\top} \hspace{-2pt} \bm{f}=\bm{0} \\ \bm{l} \leq \bm{f} \leq \bm{u} }}{\arg \min } \hspace{4pt} \bm{c}^{\top} \hspace{-2pt} \bm{f}, \\
		\text{(Dual)} & \quad \underset{\substack{\bm{B} \bm{y}+\bm{s}=\bm{c}}} {\arg \max } \sum_{i} \min \left(l_i s_i, u_i s_i\right)
	\end{split}
	\label{eq:8}
\end{equation}
where $\bm{B}_{m \times n}$, is the edge-vertex incidence matrix.
For a directed graph, the entries of the matrix $\bm{B}$ are defined as follows:
\vspace{-2pt}
\begin{equation*}
	\bm{B}_{e v} =
	\begin{cases}
		-1, & \text{if vertex } v \text{ is the tail of edge } e, \\
		1,  & \text{if vertex } v \text{ is the head of edge } e, \\
		0,  & \text{if edge } v \text{ is not incident to vertex } e.
	\end{cases}
\end{equation*}

\begin{remark}
	The maximum flow problem can be considered as a specific minimum-cost circulation problem. Here, $\bm{B}$ is the edge-vertex incidence matrix of the input graph after we added to it an edge $e(t,s)$ that connects the target $t$ to the source $s$ and its lower capacity $l_{t, s}$ be $0$ and its upper capacity $u_{t, s}$ be $\|\bm{u}\|_1$. Also, the cost vector $c$ is a vector in which $c_{t, s}=-1$ and $c_e=0$ for all other edges $e \in E$ \citep{cormen2009}.
	\label{remark:1}
\end{remark}

\subsection{Barrier Methods for Constrained Optimization} \label{sec:2:subsec3}
Consider the following optimization problem:
\begin{equation}
	f^*=\underset{\substack{\alpha(\bm{f})=\bm{0} \\ \beta(\bm{f}) \leq \bm{0} }}{\arg \min } \hspace{4pt} \xi(\bm{f})
	\label{eq:9}
\end{equation}
where $h$ represents a convex inequality constraint, $g$ represents an affine equality constraint, and $\bm{f}^*$ denote the optimal solution.

\smallskip

The interior of the constraint region is defined as $S = \left\{\bm{f} \mid \alpha(\bm{f})=0, \hspace{2pt} \beta(\bm{f})<0\right\}$. Assuming the region $S$ is nonempty and convex, we introduce the barrier function $\psi(\bm{f})$ on $S$ that is continuous and approaches infinity as $\bm{f}$ approaches to the boundary of the region, specifically $\mathlarger{\lim} _{\beta(\bm{f}) \rightarrow 0^{-}} \psi(\bm{f}) = \infty$.  One common example of barrier functions is the log barrier function, which is represented as $\log(-\beta(\bm{f}))$.

\smallskip

Given a barrier function $\psi(\bm{f})$, we can define a new objective function $\xi(\bm{f}) + \mu \hspace{1pt} \psi(\bm{f})$, where $\mu$ is a positive real number, which enables us to eliminate the inequality constraints in the original problem and obtain the following problem:
\begin{equation}
	\bm{f}^*_{\mu} = \underset{\substack{\alpha(\bm{f}) = \bm{0}}}{\arg \min } \hspace{2pt} \xi(\bm{f}) + \mu \hspace{1pt} \psi(\bm{f})
	\label{eq:10}
\end{equation}
\begin{theorem}
	For any strictly convex barrier function $\psi(\bm{f})$, convex function $\xi(\bm{f})$, and $\mu>0$, there exists a unique optimal point $\bm{f}^*_{\mu}$. Furthermore, $\lim_{\mu \to 0} \bm{f}^*_{\mu} = \bm{f}^*$, indicating that for any arbitrary $\epsilon > 0$, we can select a sufficiently small $\mu>0$ such that $\|\bm{f}^*_{\mu} - \bm{f}^*\| < \epsilon$  \citep{vandenbrand2023}. 
	\label{th:1}
\end{theorem}

	\section{Methods} \label{sec:3}

\subsection{Information Tensor} \label{sec:3:subsec:1}
In Transformer models, information propagation occurs through pathways facilitated by the attention mechanism. These pathways can be conceptualized as routes within a graph structure, where tokens are represented by nodes and computations are denoted by edges. The capacities of these edges correspond to meaningful computational quantities that reflect the flow of information through the neural network  \citep{ferrando2024, mueller2024}. 

\smallskip

First, attention weights can represent the flow of information through the neural network during the feed-forward phase of training, quantifying the importance of different input parts in generating the output \citep{abnar2020b, ferrando2024}. Additionally, the gradient of attention weights captures the flow of information during back-propagation, quantifying how changes in the output influence the attention weights throughout the network during training \citep{barkan2021}. Therefore, a combined view of attention weights and their gradients can simultaneously represent information circulation during both feed-forward and back-propagation, offering a comprehensive perspective on the network's information dynamics \citep{barkan2021, qiang2022, chefer2021a, chefer2021d}.

\smallskip

Our Generalized Attention Flow generlize this foundation by leveraging an information tensor, $\bm{\bar{A}} \! \in \! \mathbb{R}^{l \times t \times t}$, to aggregate Transformer attention weights $\bm{A}$, as defined in \cref{eq:6}. Based on the above insights, we present three aggregation functions to define information tensors \citep{barkan2021, chefer2021a}.
\begin{enumerate}
	\item \textbf{Attention Flow (AF):}\\ $\bm{\bar{A}} := \mathbb{E}_h(\bm{A})$
	\vspace{-8pt}
	\item \textbf{Attention Grad Flow (GF):}\\ $\bm{\bar{A}} := \mathbb{E}_h(\lfloor \nabla \bm{A} \rfloor_{\!+})$
	\vspace{-8pt}
	\item \textbf{Attention $\times$ Attention Grad Flow (AGF):}\\ $\bm{\bar{A}} := \mathbb{E}_h(\lfloor \bm{A} \odot \nabla \bm{A} \rfloor_{\!+})$
	\vspace{-4pt}
\end{enumerate}

Here, $\lfloor x \rfloor_{\!+} = \max(x, 0)$, $\odot$ represents the Hadamard product, $\nabla \bm{A} := \frac{\partial y_t}{\partial \bm{A}}$ where $y_t$ is the model's scalar output, and $\mathbb{E}_h$ denotes the mean across attention heads.


\begin{figure*}[ht]
	\centering
	\begin{minipage}{0.45\textwidth}
		\begin{algorithm}[H]
			\footnotesize
			\begin{algorithmic}[0]
				\Require $\bm{\bar{A}}_{l\times t \times t}$: An information tensor.
				\Ensure Tuple: $(\mathcal{A}, \bm{l}, \widetilde{\bm{l}}, \bm{u}, \widetilde{\bm{u}}, ss, st)$
				\Function{get\_backward\_capacity}{$\bm{\bar{A}}$}
				
				\vspace*{2pt}
				\State \Comment{Initialization}
				\State $l, t, \text{$\_$} \gets$ \text{$\bm{\bar{A}}$.shape()}
				\State $\text{$\beta_{\min}$} \gets \min(\bm{\bar{A}}>0)$
				\State $\text{$\beta$} \gets -\lfloor \log_{10}(\text{$\beta_{\min}$}) \rfloor$
				\State $\gamma \gets 10^\beta$
				\State $Q_{tl} \gets t \ast (l + 1) + 2$
				\State $\bm{l} \gets \text{zeros} (Q_{tl}, Q_{tl})$
				\State $\bm{u} \gets \text{zeros} (Q_{tl}, Q_{tl})$
				\State $u_{\infty} \gets t$
				
				\vspace*{2pt}
				\State \Comment{Fill super-source $\rightarrow$ First Layer}
				\For{$i$ \text{ in range}($t$)}
				\State $\bm{u}[i+1][0] \gets u_{\infty}$
				\EndFor
				
				\vspace*{2pt}
				\State \Comment{Fill Last Layer $\rightarrow$ super-target}
				\For{$i$ \text{ in range}($t$)}
				\State $\bm{u}[-1][-i-2] \gets u_{\infty}$
				\EndFor
				
				\vspace*{2pt}
				\State \Comment{Fill $j$-th Layer to $(j+1)$-th Layer}
				\For{$j$ \text{ in range}$(l)$}
				\State $\text{start} \gets t \ast j+1$
				\State $\text{mid} \gets t \ast (j+1)+1$
				\State $\text{end} \gets t \ast (j+2)+1$
				\State $\bm{u}[\text{mid:end}\ , \ \text{start:mid} ] \gets \bm{\bar{A}}_{\mathsmaller{[j, :, :]}}$
				\EndFor
				
				\vspace*{2pt}
				\State \Comment{Get Integral Version of Capacities}
				\State $\widetilde{\bm{l}} \gets\text{int}(\gamma \ast \bm{l})$
				\State $\widetilde{\bm{u}} \gets\text{int}(\gamma \ast \bm{u})$
				
				\vspace*{2pt}
				\State \Comment{Get Adjacency Matrix}
				\State $\mathcal{A} \gets \mathbb{I}_{(\bm{u}>0)}$
				
				\vspace*{2pt}
				\State \Comment{Get super-source and super-target}
				\State $ ss, st \gets t \ast (l+1) +1, 0$
				\EndFunction
			\end{algorithmic}
			\caption{Backward Information Capacity}\label{algo:1}
		\end{algorithm}
	\end{minipage}
	\hfill
	\begin{minipage}{0.45\textwidth}
		\begin{algorithm}[H]
			\footnotesize
			\begin{algorithmic}[0]
				\Require $\bm{\bar{A}}_{l\times t \times t}$: An information tensor.
				\Ensure Tuple: $( \mathcal{A},\bm{l}, \widetilde{\bm{l}}, \bm{u}, \widetilde{\bm{u}}, ss, st)$
				\Function{get\_forward\_capacity}{$\bm{\bar{A}}$}
				
				\vspace*{2pt}
				\State \Comment{Initialization}
				\State $l, t, \text{$\_$} \gets$ \text{$\bm{\bar{A}}$.shape()}
				\State $\text{$\beta_{\min}$} \gets \min(\bm{\bar{A}}>0)$
				\State $\text{$\beta$} \gets -\lfloor \log_{10}(\text{$\beta_{\min}$}) \rfloor$
				\State $\gamma \gets 10^\beta$
				\State $Q_{tl} \gets t \ast (l + 1) + 2$
				\State $\bm{l} \gets \text{zeros} (Q_{tl}, Q_{tl})$
				\State $\bm{u} \gets \text{zeros} (Q_{tl}, Q_{tl})$
				\State $u_{\infty} \gets t$
				
				\vspace*{2pt}
				\State \Comment{Fill super-source $\rightarrow$ First Layer}
				\For{$i$ \text{ in range}($t$)}
				\State $\bm{u}[0][i+1] \gets u_{\infty}$
				\EndFor
				
				\vspace*{2pt}
				\State \Comment{Fill Last Layer $\rightarrow$ super-target}
				\For{$i$ \text{ in range}($t$)}
				\State $\bm{u}[-i-2][-1] \gets u_{\infty}$
				\EndFor
				
				\vspace*{2pt}
				\State \Comment{Fill $j$-th Layer to $(j+1)$-th Layer}
				\For{$j$ \text{ in range}$(l)$}
				\State $\text{start} \gets t \ast j+1$
				\State $\text{mid} \gets t \ast (j+1)+1$
				\State $\text{end} \gets t \ast (j+2)+1$
				\State $\bm{u}[\text{start:mid} \ , \ \text{mid:end}] \gets \bm{\bar{A}}_{\mathsmaller{[j, :, :]}}^T$
				\EndFor
				
				\vspace*{2pt}
				\State \Comment{Get Integral Version of Capacities}
				\State $\widetilde{\bm{l}} \gets\text{int}(\gamma \ast \bm{l})$
				\State $\widetilde{\bm{u}} \gets\text{int}(\gamma \ast \bm{u})$
				
				\vspace*{2pt}
				\State \Comment{Get Adjacency Matrix}
				\State $\mathcal{A} \gets \mathbb{I}_{(\bm{u}>0)}$
				
				\vspace*{2pt}
				\State \Comment{Get super-source and super-target}
				\State $ ss, st \gets 0, t \ast (l+1) +1$
				\EndFunction
			\end{algorithmic}
			\caption{Forward Information Capacity}\label{algo:2}
		\end{algorithm}
	\end{minipage}
	
\end{figure*}


\subsection{Generalized Attention Flow} \label{sec:3:subsec:2}
In Generalized Attention Flow, we use the attention mechanism for feature attribution by developing a network flow representation of a Transformer or other attention-based model. We will assign capacities to the edges of this graph corresponding to information tensor defined in \cref{sec:3:subsec:1}. We then solve the maximum flow problem to compute the optimal flow passing through any output node (or, more generally, any node in any layer) to any input node. The flow traversing through an input node (token) indicates the importance or attribution of that particular node (token).

\smallskip

To determine the maximum flow from all output nodes to all input nodes, we leverage the concept of multi-commodity flow (\cref{app:1:subapp:2} and \cref{app:2}). This involves the introduction of a super-source node \(ss\) and a super-target node \(st\) with a large capacity \(u_{\infty}\). The connectivity between layers and capacities between nodes are established using the information tensors, effectively forming a layered graph (\cref{app:2}).

\smallskip

To formalize the generating of the information flow, consider a Transformer with \(l\) attention layers, an input sequence \(X \! \in  \! \mathbb{R}^{t \times d}\), and its information tensor \(\bm{\bar{A}} \! \in \! \mathbb{R}^{l \times t \times t}\). Using the information tensor $\bm{\bar{A}}$, we can construct the layered attribution graph \(\mathcal{G}\) with its adjacency matrix \(\mathcal{A}\), its edge-vertex incidence matrix \(\bm{B} \), lower capacity matrix \(\bm{l}\) and its integral version \(\widetilde{\bm{l}}\), upper capacity matrix \(\bm{u}\) and its integral version \(\widetilde{\bm{u}}\) employing either \cref{algo:1} or \cref{algo:2}. Afterward, we will substitute the obtained matrices into the primal form of \cref{eq:8} to compute the desired optimal flow.

\smallskip

To clarify \cref{algo:1} and \cref{algo:2} further, we detail the process of constructing the layered attribution graph $\mathcal{G}$, which has an adjacency matrix of shape $(2 + t \times (l + 1), 2 + t \times (l + 1))$, serving as the input for the maximum flow problem. Nodes at layer $\ell \in \{1, \dots, l\}$ and token $i \in \{1, \dots, t\}$ are designated as $v_{\ell, i}$. The following guidelines outline the process to define the upper and lower bound capacities:
\begin{itemize}
	\item To connect nodes $v_{1,i}$ to the super-target node $v_{st}$, we define $\bm{u}[0, i] = u_\infty$ for $ 1\leq i\leq t$.
	
	\item The upper-bound capacity from node $v_{\ell+1, i}$ to node $v_{\ell,j}$ is defined as $\bm{u}[\text{I}_{i, \ell+1}, \text{I}_{j, \ell}] = \bm{\bar{A}}_{\ell, i, j}$ for $\ell \! \in \! \{1, \dots ,l\}$, $i \! \in \! \{1, \dots, t\}$, and $j \! \in \! \{1, \dots, t\}$, where $\text{I}_{i, \ell+1}= i + t \ast \ell$ and $\text{I}_{j, \ell}= j + t \ast (\ell-1)$.
	
	\item To connect the super-source node $v_{ss}$ to nodes $v_{l+1,i}$, we define $\bm{u}[t \ast l +i, 1+ t \ast (l+1)] = u_\infty$ for $ 1\leq i\leq t$.
	
	\item The lower-bound capacity is defined as $\bm{l} = \bm{0}$.
\end{itemize}

\begin{figure*}[ht]
	\setcounter{subfigure}{0} 
	\centering
	\subfloat[][Schematic information flow created via \cref{algo:1}.]
	{
		\includegraphics[width=6.75cm, height=3.5cm]{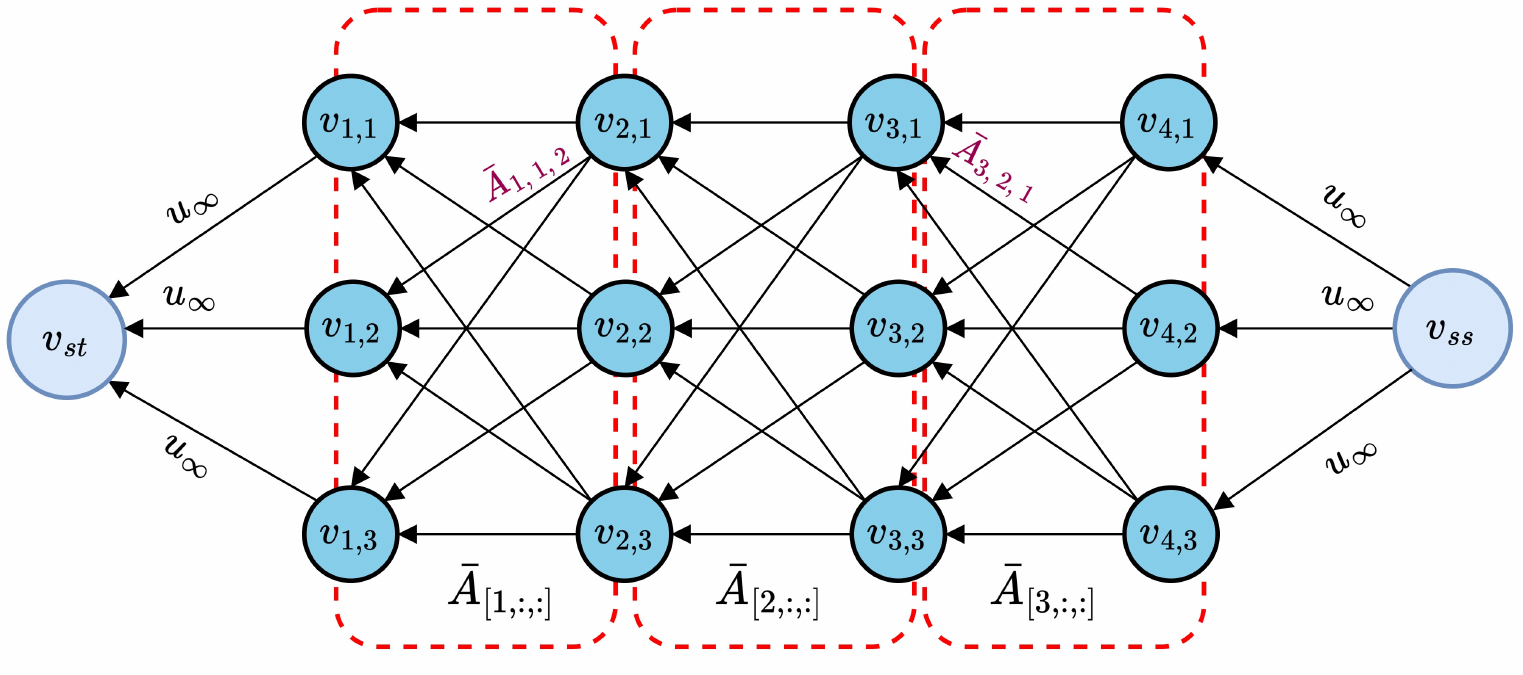}
		\label{fig1:subfig1}
	}
	\subfloat[][Schematic information flow created via \cref{algo:2}.]
	{
		\includegraphics[width=6.75cm, height=3.5cm]{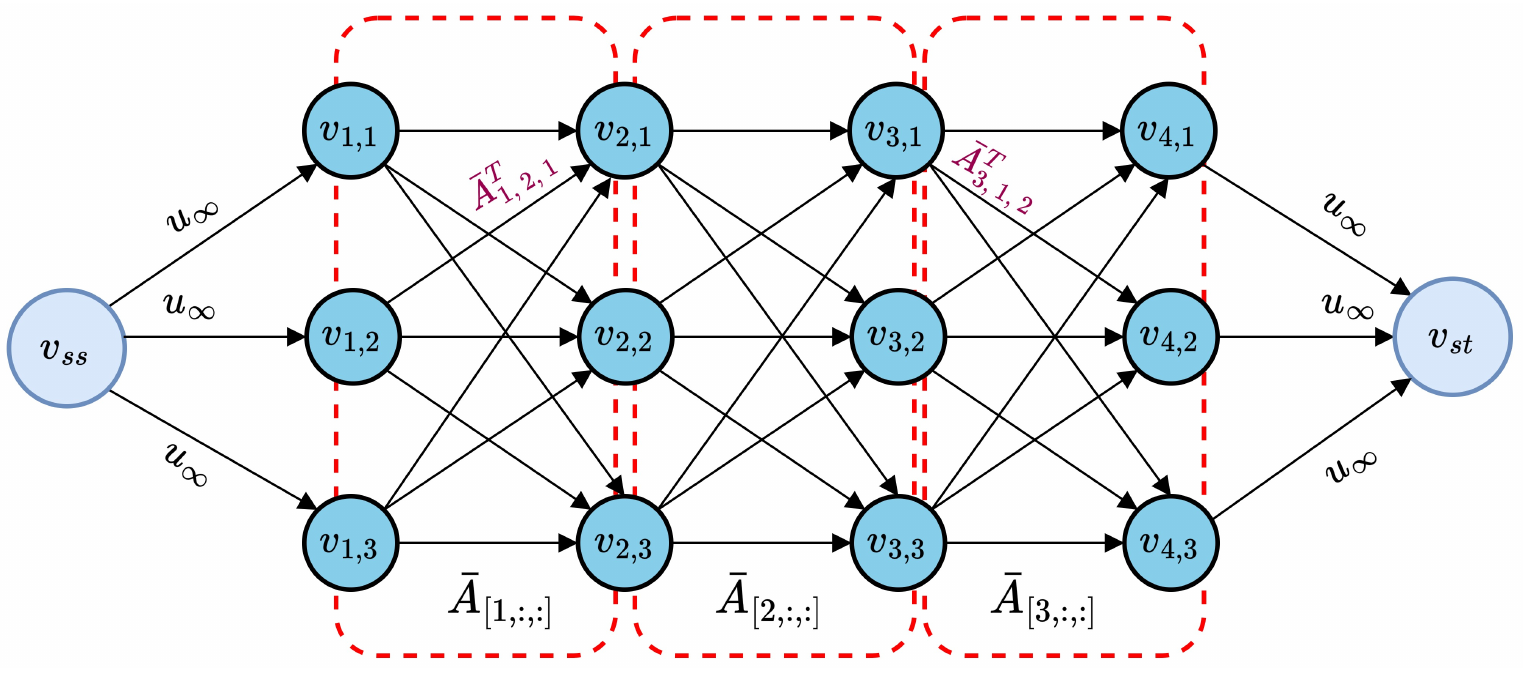}
		\label{fig1:subfig2}
	}
	\label{fig1}
	\caption{Schematics overview of Generalized Attention Flow created using \cref{algo:1} and \cref{algo:2}.}
\end{figure*}

\cref{fig1:subfig1} and \cref{fig1:subfig2} depict schematic graphs generated using the information tensor $\bm{\bar{A}} \in \mathbb{R}^{3 \times 3 \times 3}$, with \cref{algo:1} and \cref{algo:2}, respectively. While both algorithms solve the same network flow problem by constructing graphs containing a super-source and a super-target, the second algorithm differs from the first in two key aspects. First, the positions of the super-source and super-target are swapped in the second graph, such that the super-source in the first graph becomes the super-target in the second, and vice versa. Second, the direction of the edges in the second graph is reversed relative to the first.

\subsection{Non-uniqueness of Maximum Flow} \label{sec:3:subsec:3}
The maximum flow problem lacks strict convexity, meaning it does not necessarily yield the unique optimal solution. We found that the maximum flow problem associated with the graphs constructed employing Generalized Attention Flow also fails to yield the unique optimal flow (\cref{app:3}). 

\begin{observation}
	It is straightforward to verify that both \cref{algo:1} and \cref{algo:2} solve the same maximum flow problem. Therefore, determining the maximum flow in graphs generated by either \cref{algo:1} or \cref{algo:2} is equivalent and yields the same optimal value. However, it's worth noting that the optimal flows associated with them may not necessarily be equivalent, as explained in \cref{app:3}.
\end{observation}

\begin{observation}
	If two distinct feasible solutions, denoted \(f_1\) and \(f_2\), exist for a linear programming problem, then any convex combination \(\gamma_1 f_1 + \gamma_2 f_2\) forms another feasible solution. Consequently, the maximum flow problem can possess an infinite number of feasible solutions. Additionally, due to the non-uniqueness of optimal flows arising from the maximum flow problem, their projections onto any subset of nodes in the graph may also not be unique.
\end{observation}

\begin{corollary} \label{cor:1}
	Let $V$ be the set of all nodes in a layered attribution graph $\mathcal{G}(\mathcal{A}, \bm{u}, \bm{l}, \bm{c}, ss, st)$, and $N \subseteq V$, with all nodes in $N$ chosen from the same layer. Suppose $\bm{f}^*$ is the optimal solution of \cref{eq:8}, and for every $S \subseteq N$, define the payoff function $\vartheta(S) \coloneqq |\bm{f}^*(S)| = \sum_{i \in S}|f_\text{out}(i)|$, where $\left|f_\text{out}(i)\right|$ denotes the total outflow value of node $i$. Although \citet{ethayarajh2021b} claimed that for each node $i \in N$, $\phi_i(\vartheta) = |\bm{f}^*_\text{out}(i)|$ represents the Shapley value, these feature attributions are non-unique and cannot be considered Shapley values. In fact, their method for defining feature attributions is not well-defined (Proof in \cref{app:5}).
\end{corollary}

\subsection{Log Barrier Regularization of Maximum Flow} \label{sec:3:subsec:4}

To address the non-uniqueness challenge in the maximum flow problem, we reformulate the minimum-cost circulation problem as follows:
\begin{equation}
	\underset{\substack{\bm{B}^{\top} \hspace{-2pt} \bm{f}=\bm{0} \\ \beta(\bm{f}) \leq \bm{0} }}{\arg \min } \hspace{4pt} \bm{c}^{\top} \hspace{-2pt} \bm{f}
	\label{eq:11}
\end{equation}
where $\beta(\bm{f}) = (\bm{f}-\bm{l}) (\bm{f}-\bm{u})$. Consequently, the original problem can be approximated using the log barrier function as the following optimization problem:
\begin{equation}
	\begin{array}{ll}
		\underset{\substack{\bm{B}^{\top} \hspace{-2pt} \bm{f}=\bm{0}}}{\arg \min } \hspace{2pt} \bm{c}^{\top} \hspace{-2pt} \bm{f} + \psi_{\mu}(\bm{f})
	\end{array}
	\label{eq:12}
\end{equation}
where the log barrier function is:
\begin{equation}
	\begin{split}
		\psi_{\mu}(\bm{f}) &= -\mu \sum\limits_{e \in E} \log \left(-\beta(f_e)\right) \\
		&= -\mu \sum\limits_{e \in E} \left( \log \left(f_e-l_e\right) +\log \left(u_e-f_e\right) \right)
	\end{split}
	\label{eq:13}
\end{equation}
It is evident that, for any positive $\mu$ and a feasible initial solution, the barrier function guarantees that the solution derived from an iterative minimization scheme, such as interior point methods, remains feasible \citep{bubeck2015, boyd2004, madry2019}. Moreover, to obtain an $\varepsilon$-approximate solution to \cref{eq:11}, it suffices to set $\mu \leq \frac{\varepsilon}{2m}$ and solve the corresponding problem in \cref{eq:12} \citep{bubeck2015, boyd2004, madry2019}.

\smallskip

Finally, the Hessian of the objective function in \cref{eq:11} at some point $\bm{f}$ is equal to the Hessian of the barrier function, which is positive definite (assuming $\mu > 0$). This implies that the objective function is strictly convex and, consequently, \cref{eq:12} has a unique feasible solution \citep{bubeck2015, boyd2004}.

\subsection{Axioms of Feature Attributions}\label{sec:3:subsec:5}
In XAI, axioms are core principles that guide the evaluation of explanation methods, ensuring their reliability, interpretability, and fairness. These axioms provide standards to measure the effectiveness and compliance of explanation techniques. Our proposed methods meet four essential axioms, as proved by the following theorem and corollaries.

\begin{definition} [Shapley values]
	For any value function $\vartheta: 2^N \mapsto \mathbb{R}$ where $N=$ $\{1,2, \ldots, n\}$, Shapley values $\phi(\vartheta) \in \mathbb{R}^n$ can be computed by averaging the marginal contribution of each feature over all possible feature combinations:
\end{definition}
\begin{equation}
	\phi_i(\vartheta)=\mathlarger{\sum\limits_{S \subseteq N \backslash\{i\}}} \binom{n-1}{|S|}^{\! \!-1} (\vartheta(S \cup\{i\})-\vartheta(S))
	\label{eq:14}
\end{equation}

Shapley values are the unique feature attributions that satisfy four fairness-based axioms: efficiency (completeness), symmetry, linearity (additivity), and nullity \citep{shapley1952, shapley2016, young1985} (\cref{app:1:subapp:3}). Initially, a value function based on model accuracy was proposed \citep{lundberg2017b}, but since then, various alternative payoff functions have been introduced \citep{jethani2022, sundararajan2020}, each providing distinct feature importance scores.

\begin{theorem}[Log Barrier Regularization of Generalized Attention Flow Outcomes Shapley Values]
	Given a layered attribution graph $\mathcal{G}(\mathcal{A}, \bm{u}, \bm{l}, \bm{c}, ss, st)$ which has been defined using either of \cref{algo:1} or \cref{algo:2}, let $V$ be the set of all nodes in $\mathcal{G}$, and $N \subseteq V$ such that all nodes in $N$ are chosen from the same layer. Now, suppose $\bm{f}^*$ is the optimal unique solution of \cref{eq:12}, and for every $S \subseteq N$, define the payoff function $\vartheta(S) \! \coloneqq \! |\bm{f}^*(S)| = \sum_{i \in  S}|f_\text{out}(i)|$ where $\left|f_\text{out}(i)\right|$ is the total outflow value of a node $i$. Then, it can be proven that for each node $i \! \in \!  N$, $\phi_i(\vartheta)= |\bm{f}^*_\text{out}(i)|$ represents the Shapley value (Proof in \cref{app:5}).
	\label{th:2}
\end{theorem}

\begin{corollary}
	\cref{th:2} implies that the feature attributions obtained by \cref{eq:12} are Shapley values and, consequently, satisfy the axioms of \textbf{efficiency}, \textbf{symmetry}, \textbf{nullity}, and \textbf{linearity}. 
	\label{cor:2}
\end{corollary} 

\section{Experiments} \label{sec:4}
In this section, we comprehensively evaluate the effectiveness of our methods for NLP sequence classification. While our approach is versatile and applicable to various NLP tasks, including question answering and named entity recognition, which use encoder-only Transformer architectures, this assessment focuses only on sequence classification.

\subsection{Transformer Models} \label{sec:4:subsec1}
In our evaluations, we use a specific pre-trained model from the \href{https://huggingface.co/models}{HuggingFace Hub} \citep{wolf2020} for each dataset and compare our explanation methods against others to assess their performance (\cref{app:6:subapp:1}).

\subsection{Datasets} \label{sec:4:subsec2}
Our method's assessment encompasses sequence classification spanning binary classification tasks on datasets including SST2 \citep{socher2013}, Amazon Polarity \citep{mcauley2013}, Yelp Polarity \citep{zhang2016}, and IMDB \citep{maas2011}, alongside multi-class classification on the AG News dataset \citep{zhang2015}. To minimize computational overhead, we conduct the experiments using a subset of 5,000 randomly selected samples from the Amazon, Yelp, and IMDB datasets, while utilizing the full test sets for the other datasets (\cref{app:6:subapp:1}).

\subsection{Benchmark Methods} \label{sec:4:subsec3}
Our experiments compare the methods introduced in \cref{sec:3:subsec:1} with several well-known explanation methods tailored for Transformer models. To evaluate attention-based methods such as RawAtt and Rollout \citep{abnar2020b}, attention gradient-based methods like Grads, AttGrads \citep{barkan2021}, CAT, and AttCAT \citep{qiang2022}, as well as LRP-based methods such as PartialLRP \citep{voita2019a} and TransAtt \citep{chefer2021a}, we adapted the repository developed by \citet{qiang2022}. Moreover, we implemented classical attribution methods such as Integrated Gradient \citep{sundararajan2017}, KernelShap \citep{lundberg2017b}, and LIME \citep{ribeiro2016} using the Captum package \citep{kokhlikyan2020}.

\subsection{Evaluation Metric} \label{sec:4:subsec4}
\textbf{AOPC}: One of the important evaluation metrics employed is the Area Over the Perturbation Curve (AOPC), a measure that quantifies the impact of masking top $k\%$ tokens on the average change in prediction probability across all test examples. The AOPC is calculated as follows:
\begin{equation}
	\operatorname{AOPC}(k) = \frac{1}{N} \sum_{i=1}^N p\left(\hat{y} | \mathbf{x}_i\right) - p\left(\hat{y} | \tilde{\mathbf{x}}_i^k\right)
	\label{eq:15}
\end{equation}
where \(N\) is the number of examples, \(\hat{y}\) is the predicted label, \(p(\hat{y} | \cdot)\) is the probability on the predicted label, and \(\tilde{\mathbf{x}}_i^k\) is defined by masking the \(k\%\) top-scored tokens from \(\mathbf{x}_i\). To avoid arbitrary choices for \(k\), we systematically mask \(10\%, 20\%, \ldots, 90\%\) of the tokens in decreasing saliency order, resulting in \(\tilde{\mathbf{x}}_i^{10}, \tilde{\mathbf{x}}_i^{10}, \ldots, \tilde{\mathbf{x}}_i^{90}\). 

\smallskip

\textbf{LOdds}: Log-odds score is derived by averaging the difference of negative logarithmic probabilities on the predicted label over all test examples before and after masking $k \%$ top-scored tokens.
\begin{equation}
	\operatorname{LOdds}(k)=\frac{1}{N} \sum_{i=1}^N \log \frac{p\left(\hat{y} | \tilde{\mathbf{x}}_i^k\right)}{p\left(\hat{y} | \mathbf{x}_i\right)}
	\label{eq:16}
\end{equation}
\begin{table*}[ht]
	\centering
	\scriptsize
	\setlength{\tabcolsep}{4pt}
	
	\begin{tabularx}{\textwidth}{|l|*{10}{>{\centering\arraybackslash}X|}}
		\toprule
		\multicolumn{1}{l}{Methods}  & \multicolumn{2}{c}{SST2}   & \multicolumn{2}{c}{IMDB}  & \multicolumn{2}{c}{Yelp}  & \multicolumn{2}{c}{Amazon} &   \multicolumn{2}{c}{AG News} \\
		\cmidrule(l){2-3} \cmidrule(l){4-5} \cmidrule(l){6-7} \cmidrule(l){8-9} \cmidrule(l){10-11} 
		\multicolumn{1}{l}{}  & \multicolumn{1}{c}{AOPC$\uparrow$}  & \multicolumn{1}{c}{LOdds$\downarrow$}  & \multicolumn{1}{c}{AOPC$\uparrow$}  & \multicolumn{1}{c}{LOdds$\downarrow$}  & \multicolumn{1}{c}{AOPC$\uparrow$}  & \multicolumn{1}{c}{LOdds$\downarrow$} & \multicolumn{1}{c}{AOPC$\uparrow$}  & \multicolumn{1}{c}{LOdds$\downarrow$}  & \multicolumn{1}{c}{AOPC$\uparrow$}  & \multicolumn{1}{c}{LOdds$\downarrow$} \\
		\hline
		\rowcolor{palegrey} RawAtt      & 0.348 & -0.973      & 0.329 & -1.393     & 0.383 & -1.985      & 0.353 & -1.593     & 0.301 & -1.105   \\
		\rowcolor{palegrey} Rollout     & 0.322 & -0.887      & 0.354 & -1.456     & 0.260 & -0.987      & 0.304 & -1.326     & 0.249 & -0.983   \\
		\hline
		\rowcolor{palegrey} Grads       & 0.354 & -0.313      & 0.324 & -1.271     & 0.412 & -1.994      & 0.405 & -1.793     & 0.327 & -1.319   \\
		\rowcolor{palegrey} AttGrads    & 0.367 & -0.654      & 0.337 & -1.226     & 0.423 & -1.978      & 0.419 & -1.918     & 0.348 & -1.477   \\
		\rowcolor{palegrey} CAT         & 0.369 & -1.175      & 0.332 & -1.274     & 0.417 & -1.992      & 0.381 & -1.639     & 0.325 & -1.226  \\
		\rowcolor{palegrey} AttCAT      & 0.405 & -1.402      & 0.371 & -1.642     & 0.431 & -$\bm{2.134}$      & 0.427 & -2.041     & 0.387 & -1.688  \\
		\hline
		\rowcolor{palegrey} PartialLRP  & 0.371 & -1.171      & 0.323 & -1.321     & $\bm{0.443}$ & -2.018      & 0.384 & -1.945     & 0.356 & -1.627   \\
		\rowcolor{palegrey} TransAtt    & 0.399 & -1.286      & 0.355 & -1.513     & 0.411 & -1.473      & 0.375 & -1.875     & 0.377 & -1.318   \\
		\hline
		\rowcolor{palegrey} LIME        & 0.362 & -1.056      & 0.347 & -1.379     & 0.361 & -1.568      & 0.358 & -1.612     & 0.349 & -1.538   \\
		\rowcolor{palegrey} KernelShap  & 0.382 & -1.259      & 0.367 & -1.423     & 0.385 & -1.736      & 0.374 & -1.717     & 0.351 & -1.413   \\
		\rowcolor{palegrey} IG          & 0.401 & -1.205      & 0.350 & -1.443     & 0.409 & -1.924      & 0.434 & -2.024     & 0.393 & -1.681    \\
		\hline
		\rowcolor{palegrey} AF          & 0.371 & -1.215      & 0.313 & -1.297     & 0.398 & -1.886      & 0.388 & -1.923     & 0.352 & -1.282   \\
		\rowcolor{palegrey} GF          & 0.412 & -1.616      & 0.491 & -1.718     & 0.396 & -1.654      & 0.421 & -2.006     & 0.366 & -1.513   \\
		\rowcolor{palegrey} AGF         & $\bm{0.427}$ & -$\bm{1.687}$      & $\bm{0.498}$ & -$\bm{1.849}$     & 0.429 & -1.982      & $\bm{0.439}$ & -$\bm{2.103}$     & $\bm{0.398}$ & -$\bm{1.693}$    \\ \hline
	\end{tabularx}
	\caption{AOPC and LOdds scores of all methods in explaining the Transformer-based model across datasets when we mask \textbf{top} $k\%$ tokens. Higher AOPC and lower LOdds are desirable, indicating a strong ability to mark important tokens. Best results are in bold, and differences between AGF and benchmarks are statistically significant according to the ASO test (\cref{app:3:subapp:3}).}
	\label{tab:1}
	
	\medskip
	
	\begin{tabularx}{\textwidth}{|l|*{10}{>{\centering\arraybackslash}X|}}
		\toprule
		\multicolumn{1}{l}{Methods}  & \multicolumn{2}{c}{SST2}   & \multicolumn{2}{c}{IMDB}  & \multicolumn{2}{c}{Yelp}  & \multicolumn{2}{c}{Amazon} &   \multicolumn{2}{c}{AG News} \\
		\cmidrule(l){2-3} \cmidrule(l){4-5} \cmidrule(l){6-7} \cmidrule(l){8-9} \cmidrule(l){10-11} 
		\multicolumn{1}{l}{}  & \multicolumn{1}{c}{AOPC$\downarrow$}  & \multicolumn{1}{c}{LOdds$\uparrow$}  & \multicolumn{1}{c}{AOPC$\downarrow$}  & \multicolumn{1}{c}{LOdds$\uparrow$}  & \multicolumn{1}{c}{AOPC$\downarrow$}  & \multicolumn{1}{c}{LOdds$\uparrow$} & \multicolumn{1}{c}{AOPC$\downarrow$}  & \multicolumn{1}{c}{LOdds$\uparrow$}  & \multicolumn{1}{c}{AOPC$\downarrow$}  & \multicolumn{1}{c}{LOdds$\uparrow$} \\
		\hline
		\rowcolor{palegrey} RawAtt      & 0.184 & -0.693      & 0.151 & -0.471           & 0.157 & -0.747      & 0.129 & -0.281     & 0.101 & -0.427   \\
		\rowcolor{palegrey} Rollout     & 0.221 & -0.773      & 0.123 & -0.425           & 0.169 & -0.734      & 0.171 & -0.368    & 0.117 & -0.471   \\
		\hline
		\rowcolor{palegrey} Grads       & 0.234 & -0.776      & 0.083 & -0.203           & 0.131 & -0.641     & 0.134 & -0.254     & 0.083 & -0.390   \\
		\rowcolor{palegrey} AttGrads    & 0.217 & -0.713      & 0.088 & -0.243           & 0.127 & -0.603     & 0.135 & -0.266     & 0.071 & -0.351   \\
		\rowcolor{palegrey} CAT         & 0.247 & -0.874      & 0.099 & -0.327           & 0.134 & -0.659      & 0.126 & -0.240     & 0.104 & -0.419 \\
		\rowcolor{palegrey} AttCAT      & 0.143 & -0.412     & 0.041 & -0.092            & $\bm{0.103}$ & -$\bm{0.339}$      & 0.115 & -0.148     & 0.057 & -0.219  \\
		\hline
		\rowcolor{palegrey} PartialLRP & 0.163 & -0.527      & 0.057 & -0.116     & 0.116 & -0.486      & 0.167 & -0.327     & 0.056 & -0.204   \\
		\rowcolor{palegrey} TransAtt   & 0.148 & -0.483      & 0.045 & -0.107     & 0.123 & -0.538     & 0.113 & -0.140     & 0.049 & -0.173   \\
		\hline
		\rowcolor{palegrey} LIME   & 0.173 & -0.603     & 0.076 & -0.141     & 0.143 & -0.687     & 0.158 & -0.263     & 0.075 & -0.372   \\
		\rowcolor{palegrey} KernelShap & 0.197 & -0.729  & 0.039 & -0.084     & 0.135 & -0.645      & 0.174 & -0.351     & 0.067 & -0.219   \\
		\rowcolor{palegrey} IG     & 0.150 & -0.532      & 0.026 & -0.064     & 0.130 & -0.617      & 0.134 & -0.241     & 0.052 & -0.191    \\
		\hline
		\rowcolor{palegrey} AF       & 0.199 & -0.747       & 0.061 & -0.148    & 0.153 & -0.689      & 0.388 & -1.923     & 0.106 & -0.402   \\
		\rowcolor{palegrey} GF       & 0.154 & -0.497       & 0.034 & -0.079     & 0.149 & -0.654      & 0.130 & -0.267     & 0.090 & -0.313   \\
		\rowcolor{palegrey} AGF      & $\bm{0.084}$ & -$\bm{0.263}$          & $\bm{0.014}$ & -$\bm{0.039}$    & 0.121 & -0.504      & $\bm{0.092}$ & -$\bm{0.114}$     & $\bm{0.037}$ & -$\bm{0.134}$    \\ \hline
	\end{tabularx}
	\caption{AOPC and LOdds scores of all methods in explaining the Transformer-based model across datasets when we mask \textbf{bottom} $k\%$ tokens. Lower AOPC and higher LOdds are desirable, indicating a strong ability to mark important tokens. Best results are in bold, and differences between AGF and benchmarks are statistically significant according to the ASO test (\cref{app:3:subapp:3}).}
	\label{tab:2}
\end{table*}

\section{Results} \label{sec:5}
We assessed various explanation methods by masking the top $k\%$ of tokens across multiple datasets and measuring their AOPC and LOdds scores. \Cref{tab:1} presents the average scores for different $k$ values, proving that the AGF method consistently outperforms others by achieving the highest AOPC and lowest LOdds scores, effectively identifying and masking the most important tokens for model predictions. Additionally, the GF method surpasses most baseline approaches. Evaluations based on classification metrics further confirm these findings. (\cref{app:3:subapp:1}, \cref{app:3:subapp:2}).

\smallskip

Additionally, we assessed the aforementioned explanation methods by masking the bottom $k\%$ of tokens across datasets and measuring AOPC and LOdds scores, detailed in \cref{tab:2}. The AGF method achieved the highest LOdds and lowest AOPC across most datasets, highlighting its ability to pinpoint important tokens for model predictions, with the GF method also surpassing many baseline methods in this context.

\smallskip

In contrast, the Yelp dataset presents a unique challenge, as our methods do not perform optimally in terms of AOPC and LOdds metrics. This is likely due to the prevalence of conversational language, slang, and typos in Yelp reviews, which adversely affect the AGF method's performance more than others.

\section{Conclusion} \label{sec:6}
In this study, we propose Generalized Attention Flow, an extension of Attention Flow. The core idea behind Generalized Attention Flow is applying the log barrier method to the maximum flow problem, defined by information tensors, to derive feature attributions. By leveraging the log barrier method, we resolve the non-uniqueness issue in optimal flows originating from the maximum flow problem, ensuring that our feature attributions are Shapley values and satisfy efficiency, symmetry, nullity, and linearity axioms.

\smallskip

Our experiments across various datasets indicate that our proposed AGF (and GF) method generally outperforms other feature attribution methods in most evaluation scenarios. It could be valuable for future research to explore whether alternative definitions of the information tensor could enhance AGF's effectiveness.

\section*{Limitations} \label{sec:7}
The primary limitation of our proposed method is the increased running time of the optimization problem in \cref{eq:12} as the number of tokens grows \citep{lee2020a, brand2021a}. Moreover, it's important to note that optimization problems generally cannot be solved in parallel. 

\smallskip

Although recent theoretical advancements have developed almost-linear time algorithms to solve the optimization problem described in \cref{eq:12} (\cref{tab:7}), their computational cost is still significant, particularly when we have long input sequences. Nevertheless, we found that the practical runtime of our method is comparable to other XAI methods (\cref{tab:8}).


\bibliographystyle{acl_natbib}
\bibliography{neurips2024_conference}
\pagebreak

\appendix

\section{Preliminaries} \label{app:1}

\subsection{Maximum Flow} \label{app:1:subapp:1}
\begin{definition}[Network Flow]
	Given a network $G=(V, E, s, t, \bm{u})$, where $s$ and $t$ are the source and target nodes respectively and $u_{i j}$ is the capacity for the edge $(i,j)\! \in \!  E$, a flow is characterized as a function $ f: E \rightarrow \mathbb{R}^{\geq 0} $ s.t.
	\begin{equation}
		\begin{array}{ll}
			f_{i j} \leq u_{i j} \quad \forall(i, j) \! \in \!  E \hspace{8pt} \\
			\mathlarger{\mathlarger{\sum\limits_{j:(i, j) \in  E}}} f_{i j} - \hspace{-3pt} \mathlarger{\mathlarger{\sum\limits_{j:(j, i)  \in  E}}} f_{j i} = 0, \quad \forall i \in  V, i \neq s, t \hspace{8pt}
		\end{array}
		\label{eq:17}
	\end{equation}
\end{definition}

We define $\left|f_\text{out}(i)\right|$ to be the total outflow value of a node $i$ and $\left|f_\text{in}(i)\right|$ to be the total inflow value of a node $i$. For a given set $K \subseteq V$ of nodes, we define $|f(K)|= \sum_{i \in  K}|f_\text{out}(i)|$ for every flow $f$. The value of a flow in a given network $ G = (V, E, s, t, \bm{u}) $ is denoted as $ |f| = \sum_{v:(s, v)} f_{s v} - \sum_{v:(v, s)} f_{v s} = |f_\text{out}(s)| - |f_\text{in}(s)|$, and a maximum flow is identified as a feasible flow with the highest attainable value.

\subsection{Multi-Commodity Maximum Flow} \label{app:1:subapp:2} 
The multi-commodity maximum flow problem aims to generalize the maximum flow problem by considering multiple source-sink pairs instead of a single pair. The objective of this new problem is to determine multiple optimal flows, $ f^1(\cdot, \cdot), \ldots, f^r(\cdot, \cdot) $, where each $f^k(\cdot, \cdot) $ represents a feasible flow from source $ s_k $ to sink $ t_k $

\begin{equation}
	\footnotesize
	\sum_{k=1}^r f^k(i, j) \leq u(i, j) \quad \forall(i, j) \! \in \!  E
	\label{eq:18}
\end{equation}
Therefore, the multi-commodity maximum flow problem aims to maximize the objective function $ \sum_{k=1}^r \sum_{v:(v, s_k)} f^k(s_k, v) $.

\smallskip

To solve the multi-commodity maximum flow problem, we can easily transform it into a standard maximum flow problem. This can be achieved by introducing two new nodes, a "super-source" node $ss$ and a "super-target" node $st$. The "super-source" node $ss$ should be connected to all the original sources $s_i$ through edges with finite capacities, while the "super-target" node $st$ should be connected to all the original sinks $t_i$ through edges with finite capacities:

\begin{itemize}
	\item Each outgoing edge from the "super-source" node $ss$ to each source node $s_i$ is assigned a capacity that is equal to the total capacity of the outgoing edges from the source node $s_i$.
	
	\item Each incoming edge from an original "super-target" node $st$ to each sink node $t_i$ is assigned a capacity that is equal to the total capacity of the incoming edges to the sink node $t_i$.
\end{itemize}

It is easy to demonstrate that the maximum flow from $ss$ to $st$ is equivalent to the maximum sum of flows in a feasible multi-commodity flow within the original network.

\subsection{Shapley values} \label{app:1:subapp:3}

The Shapley value, introduced by \citet{shapley1952}, concerns the cooperative game in the coalitional form $(N, \vartheta)$, where $N$ is a set of $n$ players and $\vartheta: 2^N \rightarrow \mathbb{R}$ with $\vartheta(\emptyset)=0$ is the value (payoff) function. In the game, the marginal contribution of the player $i$ to any coalition $S$ with $i \! \notin \! S$ is considered as $\vartheta(S \cup i)-\vartheta(S)$. These Shapley values are the only constructs that jointly satisfy the efficiency, symmetry, nullity, and additivity axioms \citep{shapley1952, young1985}:

\smallskip

\textbf{Efficiency:} The Shapley values must add up to the total value of the game, which means $\sum_{i \! \in \!  N} \phi_i(\vartheta) = \vartheta(N)$.

\smallskip

\textbf{Symmetry:} If two players are equal in their contributions to any coalition, they should receive the same Shapley value. Mathematically, if $\vartheta(S \cup \{i\}) = \vartheta(S \cup \{j\})$ for all $S \subseteq N \backslash \{i, j\}$, then $\phi_i(\vartheta) = \phi_j(\vartheta)$.

\smallskip

\textbf{Nullity (Dummy):} If a player has no impact on any coalition, their Shapley value should be zero. Mathematically, if $\vartheta(S \cup \{i\}) = \vartheta(S)$ for all $S \subseteq N \backslash \{i\}$, then $\phi_i(\vartheta) = 0$.

\smallskip

\textbf{Linearity:} If the game $\vartheta(\cdot)$ is a linear combination of two games $\vartheta_1(\cdot), \vartheta_2(\cdot)$ for all $S \subseteq N$, i.e. $\vartheta(S)=$ $\vartheta_1(S)+\vartheta_2(S)$ and $(c \cdot \vartheta)(S)=c \cdot \vartheta(S), \forall c \! \in \!  \mathbb{R}$, then the Shapley value in the game $\vartheta$ is also a linear combination of that in the games $\vartheta_1$ and $\vartheta_2$, i.e. $\forall i \! \in \!  N, \phi_i(\vartheta_1)=\phi_i(\vartheta_1)+\phi_i(\vartheta_2)$ and $\phi_i(c \cdot \vartheta)=c \cdot \phi_i(\vartheta)$.

Considering these axioms, the attribution of a player $j$ is uniquely given by \citep{shapley1952, young1985}:
\begin{equation}
	\phi_i(\vartheta)=\mathlarger{\sum\limits_{S \subseteq N \backslash\{i\}}} \binom{n-1}{|S|}^{\! \!-1} (\vartheta(S \cup\{i\})-\vartheta(S))
	\label{eq:19}
\end{equation}

The difference term   $ \vartheta(S \cup \{j\}) - \vartheta(S) $ represents the $i$-th feature’s contribution to the subset $S$, while the summation provides a weighted average across all subsets excluding $i$.  

%

Initially, a value (payoff) function based on model accuracy was suggested \citep{lundberg2017b}. Since then, various alternative coalition functions have been proposed \citep{jethani2022, sundararajan2020}, each resulting in a different feature importance score. Many of these alternative approaches are widely used and have been shown to outperform the basic SHAP method \citep{lundberg2017b} in empirical studies.

\begin{figure*}[!htbp]
	\setcounter{subfigure}{0} 
	\centering
	\subfloat[][Network flow with multiple sources and targets.]
	{\includegraphics[width=5.00cm, height=3.25cm]{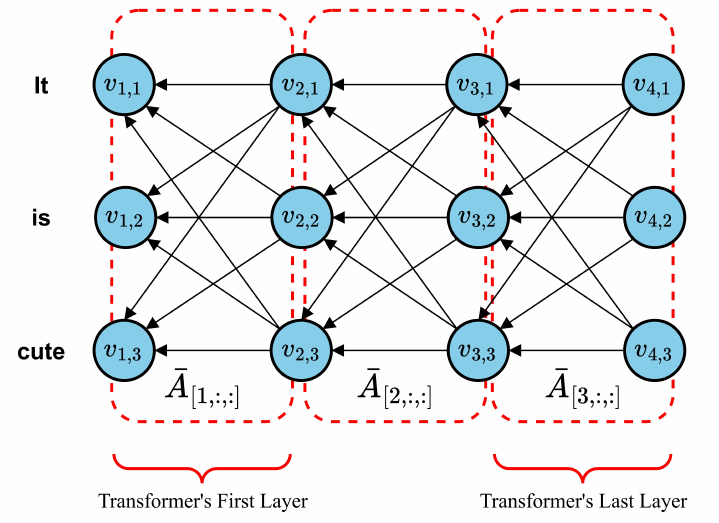}
		\label{fig2:subfig1}}
	\quad 
	\centering
	\subfloat[][Network flow including super-source and super-target.]
	{\includegraphics[width=6.75cm, height=3.25cm]{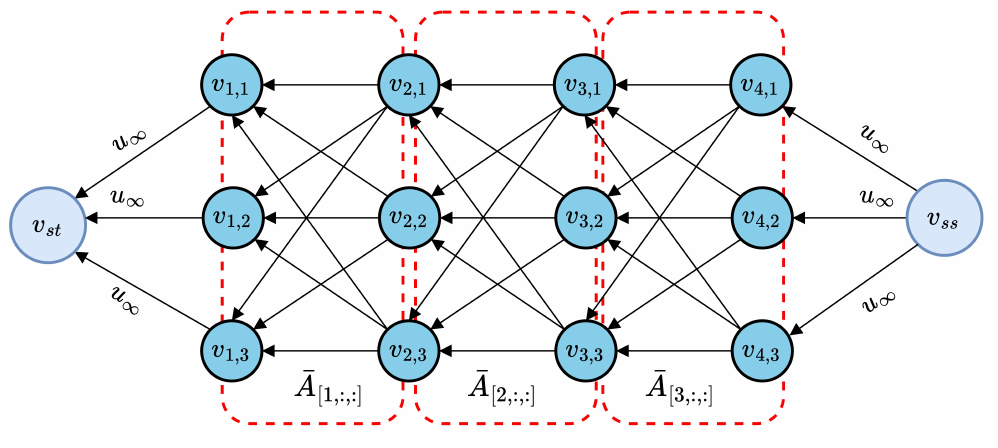}
		\label{fig2:subfig2}}
	
	\centering
	\subfloat[][Network flow defined for MCC problem.]
	{\includegraphics[width=10cm, height=3.30cm]{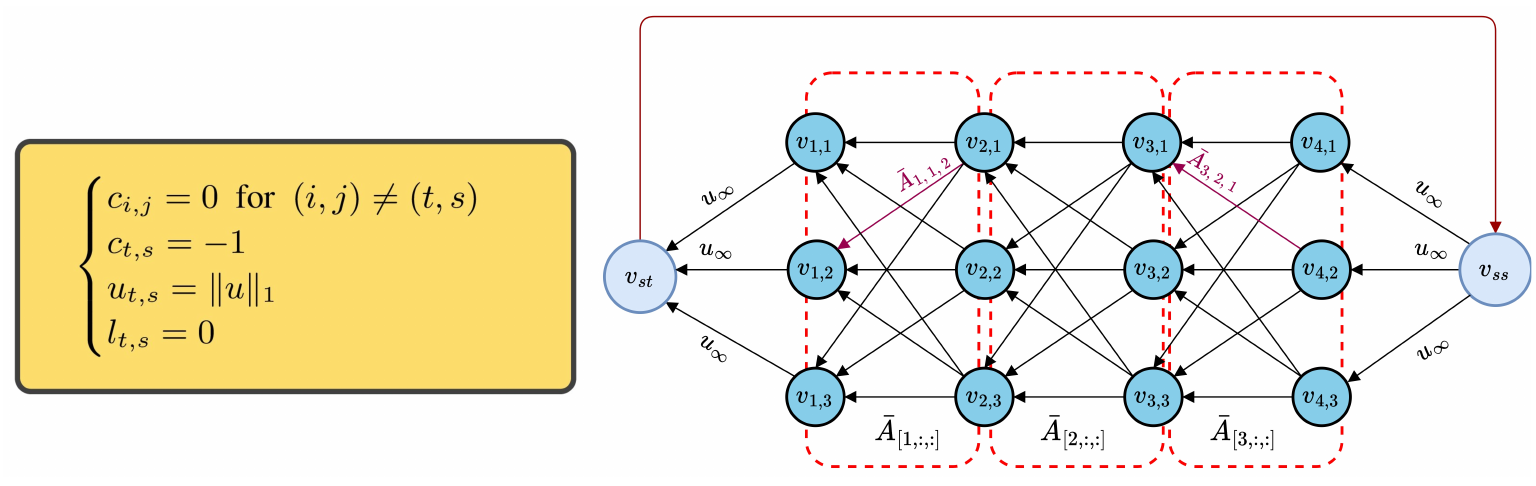}
		\label{fig2:subfig3}}
	
	\caption{Initial network flow to be used in our proposed method, the multi-commodity flow with multiple sources and targets, and the network flow for MCC problems.}
	\label{fig2}
	
\end{figure*}

\section{Network Flow Generation}\label{app:2}
\cref{fig2} will detail the process of defining a graph network and its parameters in our proposed method using \cref{algo:1}. It is worth noting that the same procedure can be defined using \cref{algo:2}. To solve the maximum flow or MCC problem within this graph network, we should compute network flow with multiple sources and targets, assigning all nodes in the first and last layers of transformers as sources and targets, respectively (\cref{fig2:subfig1}).

\smallskip

To solve this problem, we utilize the concept of multi-commodity maximum flow (multiple-sources multiple-targets maximum flow) by introducing a super-source node $ss$ and a super-target node $st$ (\cref{fig2:subfig2}). To define the upper-bound and lower-bound capacities of this new graph network, we utilize the procedure defined in \cref{sec:3:subsec:2}. In the last step, we add a new edge from the super-target node $st$ to the super-source node $ss$ and compute the cost vector, upper-bound capacities, and lower-band capacities according to \cref{fig2:subfig3}. Finally, we can input all derived parameters into \cref{eq:12}, solve the optimization problem, and calculate feature attributions.

\section{Non-uniqueness of Maximum Flow} \label{app:3}
\begin{figure*}[!htbp]
	\centering
	\includegraphics[width=13.75cm, height=10cm]{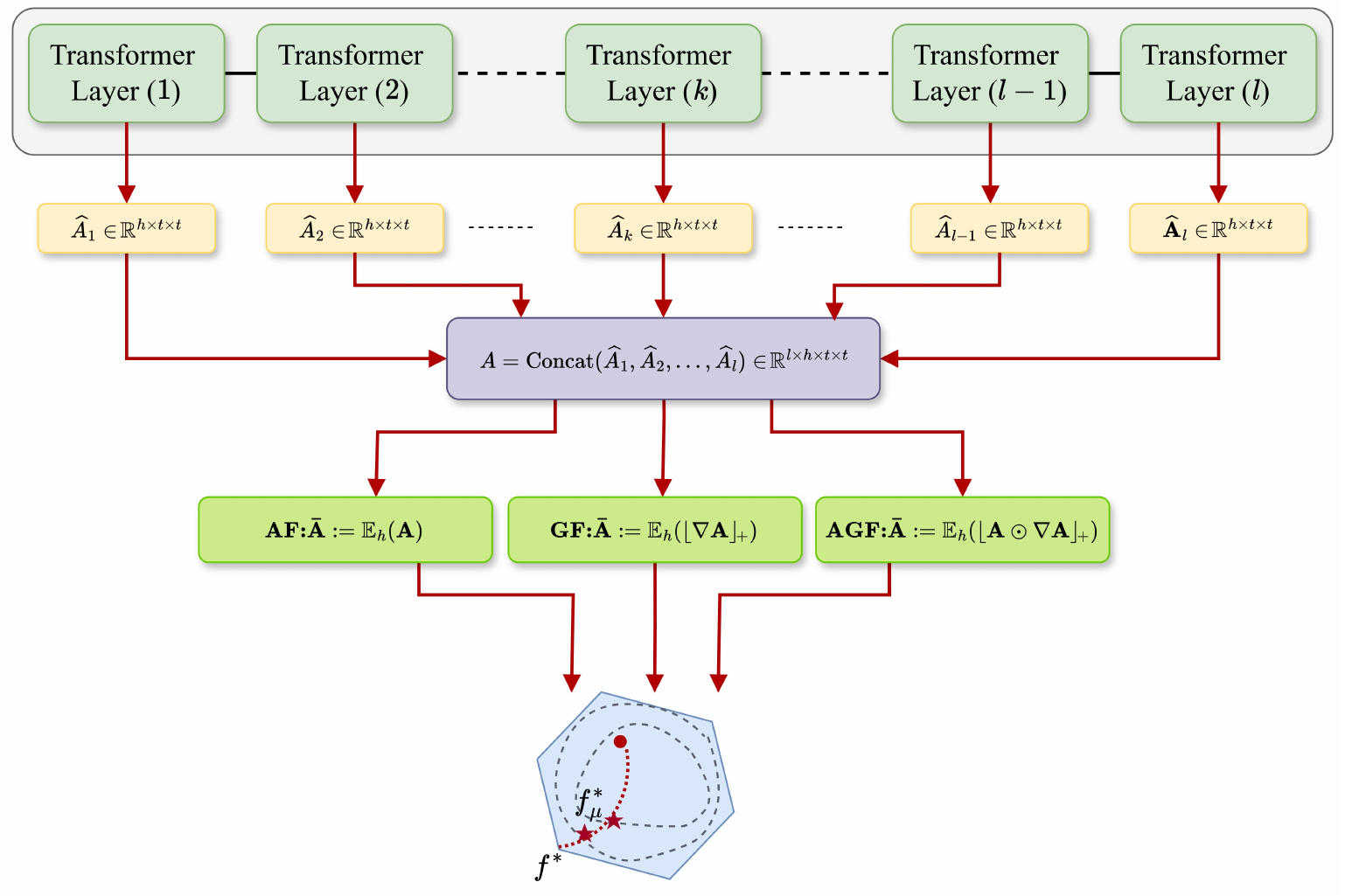}
	\caption{Overview of how the proposed method computes the unique optimal flow using the log barrier method, attention weights, and their gradients in Transformers.}
	\label{fig3}
\end{figure*}

\cref{fig3} visually describes our proposed approach for computing feature attributions. Using maximum flow to derive these attributions produces a convex set containing all optimal flows, which makes it unsuitable as a feature attribution technique. In contrast, our proposed approach, which utilizes the log barrier method, generate a unique optimal flow and provides an interpretable set of feature attributions.

\smallskip

\begin{figure*}[!htbp]
	\setcounter{subfigure}{0} 
	\subfloat[][ Network flow generated via \cref{algo:1}]
	{\includegraphics[width=6.75cm, height=5.65cm]{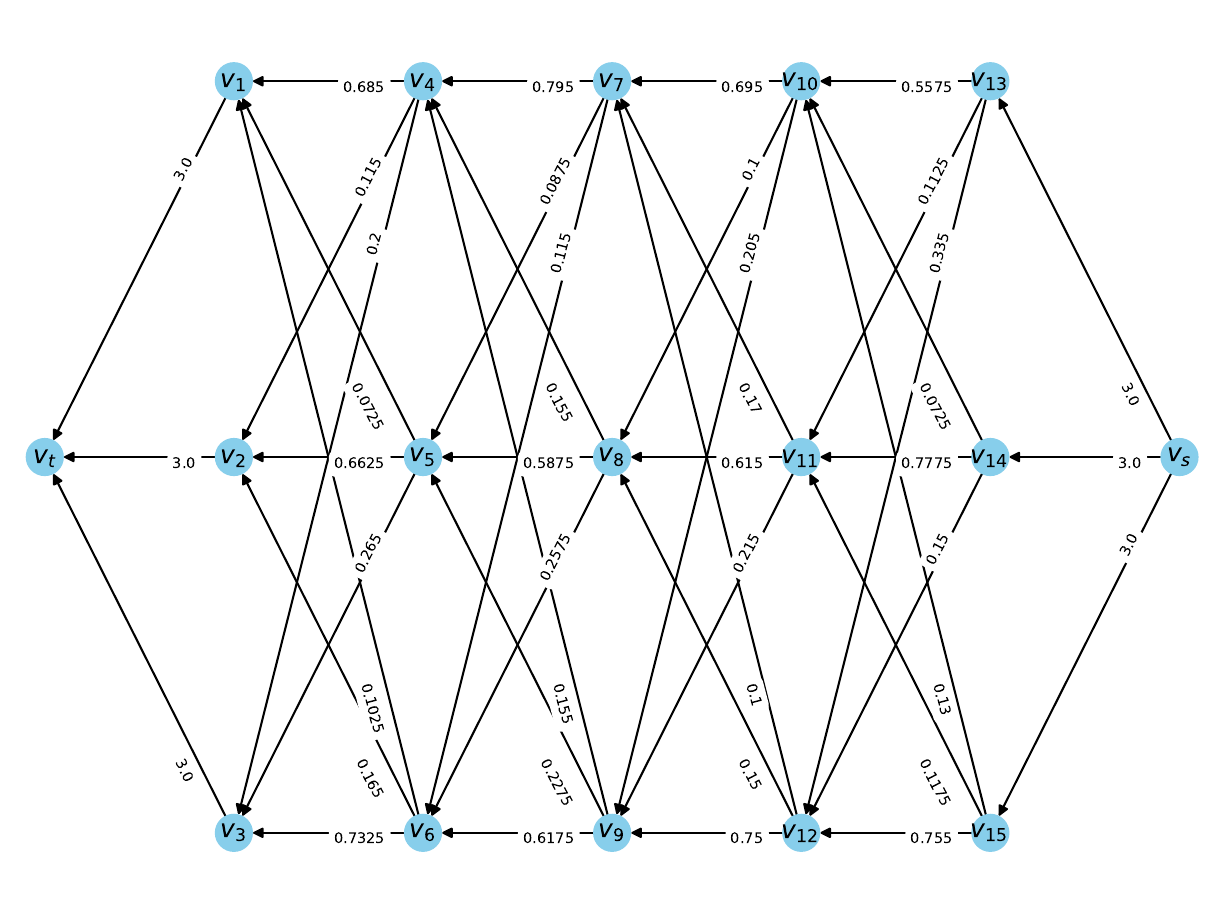}
		\label{fig4:subfig1}}
	\qquad 
	\subfloat[][Network flow generated via \cref{algo:2}]
	{\includegraphics[width=6.75cm, height=5.65cm]{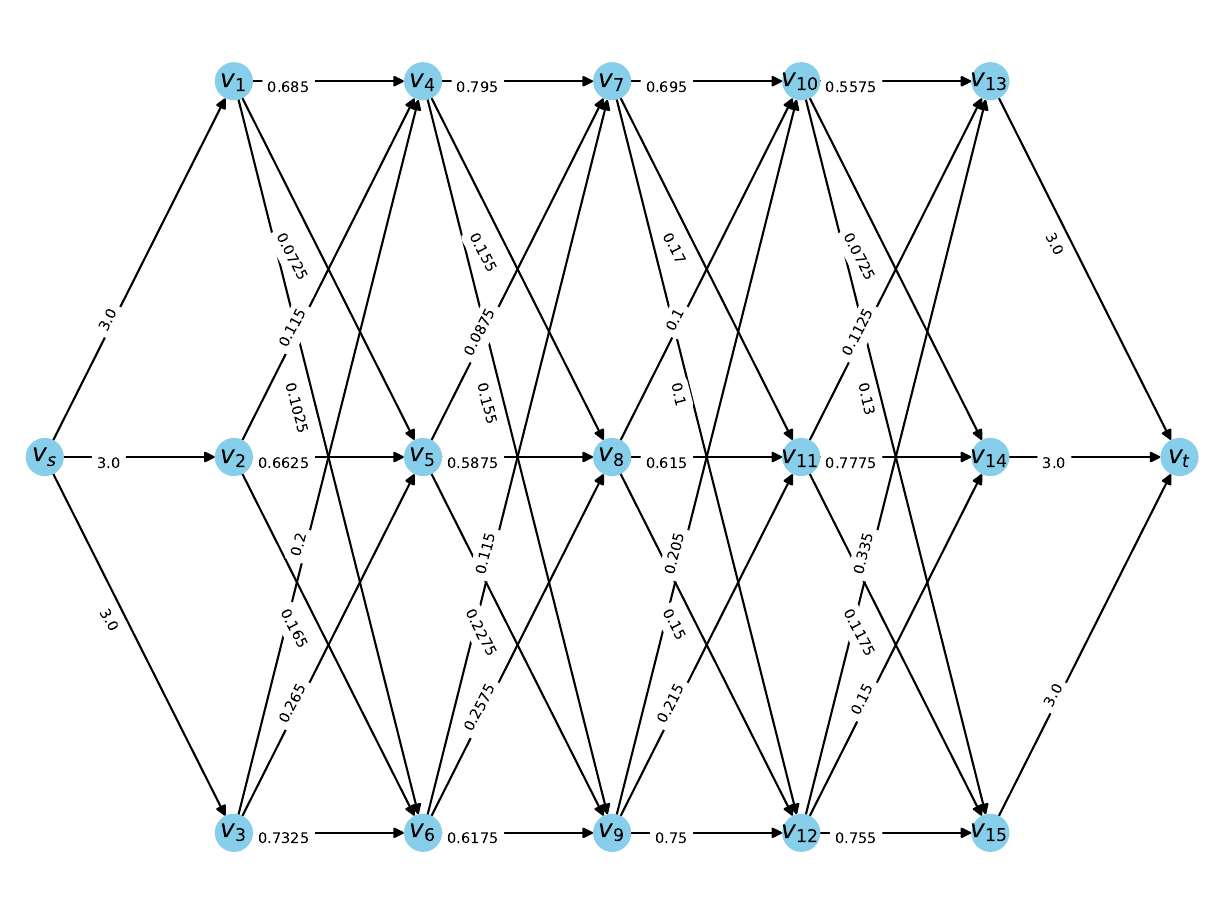}
		\label{fig4:subfig2}}
	\\
	
	\subfloat[][Optimal flow generated via \cref{algo:1}]
	{\includegraphics[width=6.75cm, height=5.65cm]{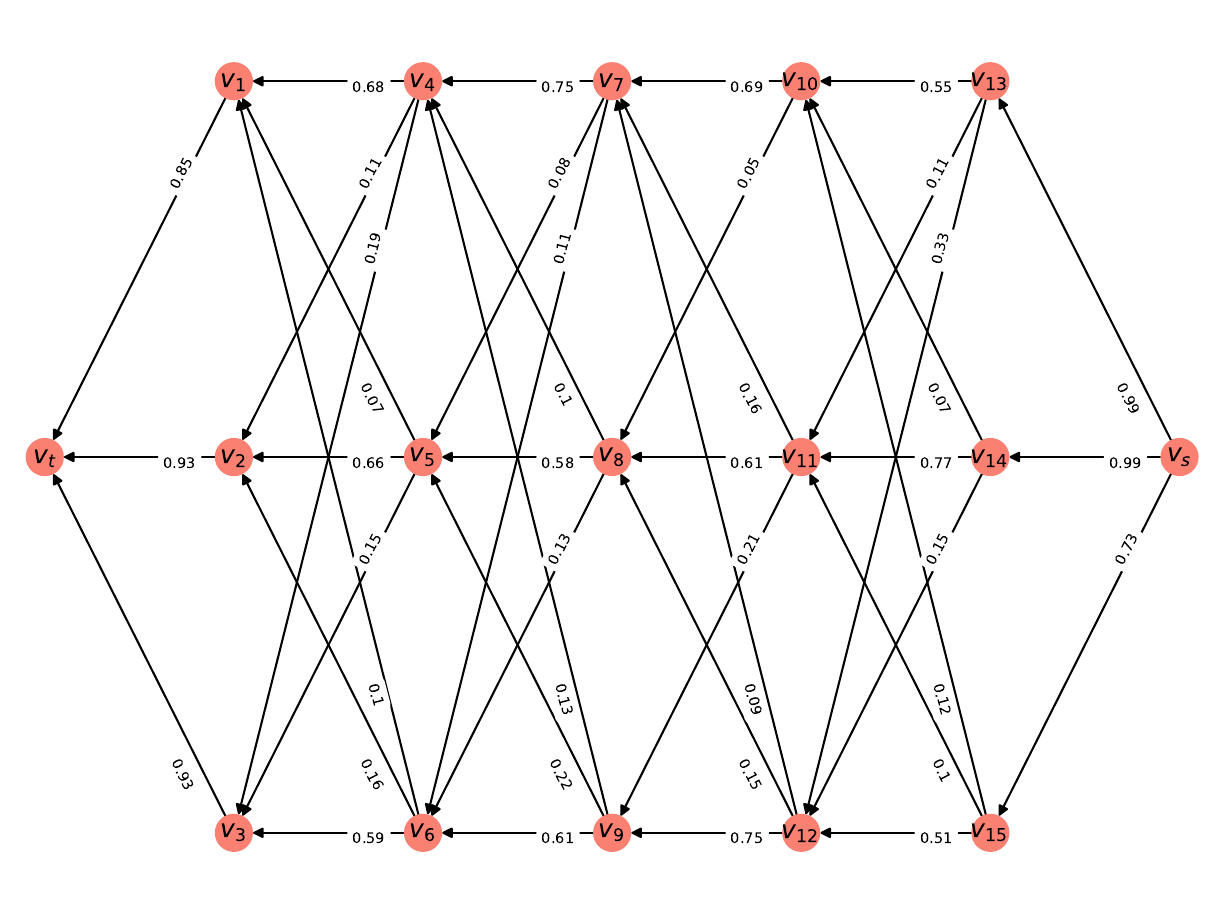}
		\label{fig4:subfig3}}
	\qquad
	\subfloat[][ Optimal flow generated via \cref{algo:2}]
	{\includegraphics[width=6.75cm, height=5.65cm]{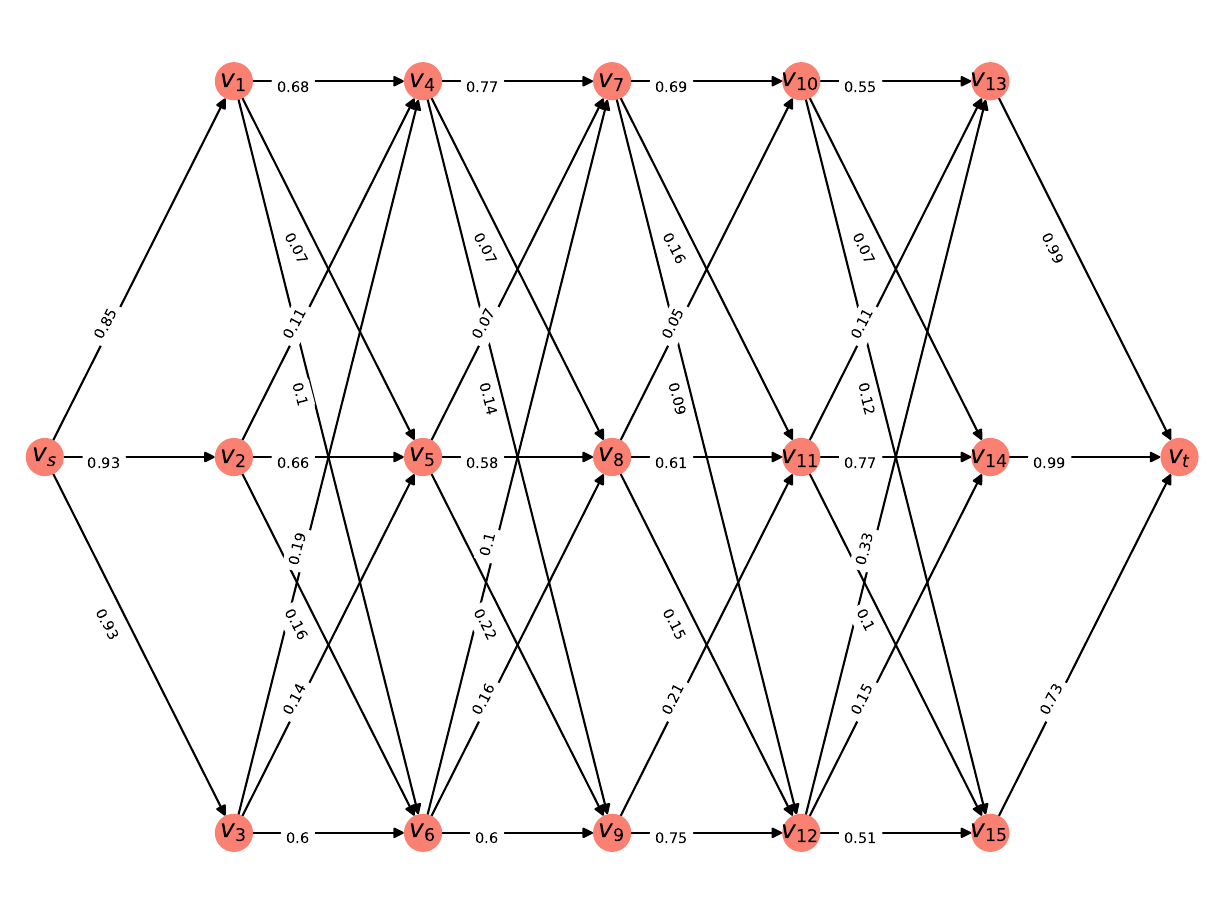}
		\label{fig4:subfig4}}
	%
	
	\caption{Network flows and optimal flows generated by \cref{algo:1} and \cref{algo:2}. The optimal flows computed using \cref{algo:1} and \cref{algo:2} are not equivalent.}
	\label{fig4}
	
\end{figure*}

\cref{fig4} depicts the capacities and optimal flows obtained by solving maximum flow problem on the network, defined with the synthetic information tensor $\bm{\bar{A}} \! \in \! \mathbb{R}^{4 \times 3 \times 3}$ as input, using \cref{algo:1} and \cref{algo:2}. While the optimal values are the same for both algorithms, their optimal flows differ. Notably, significant differences in flows between node pairs \textcolor{usflag}{$\{\bm{v_4}, \bm{v_8}\}$} and \textcolor{usflag}{$\{\bm{v_3}, \bm{v_5}\}$} are visible in \cref{fig4:subfig3} and \cref{fig4:subfig4}. Additionally, we evaluated the optimal value and its optimal flow generated by both algorithms across various combinations of token numbers $t$ and Transformer layers $l$. Our findings indicate that the optimal flows from the two algorithms do not coincide in any scenario.

\begin{figure*}[!htbp]
	\setcounter{subfigure}{0} 
	\centering
	\subfloat[][Normalized feature attributions for Transformer's input layer generated by \cref{algo:1} for different information tensors.]
	{
		\includegraphics[width=15cm, height=3cm]{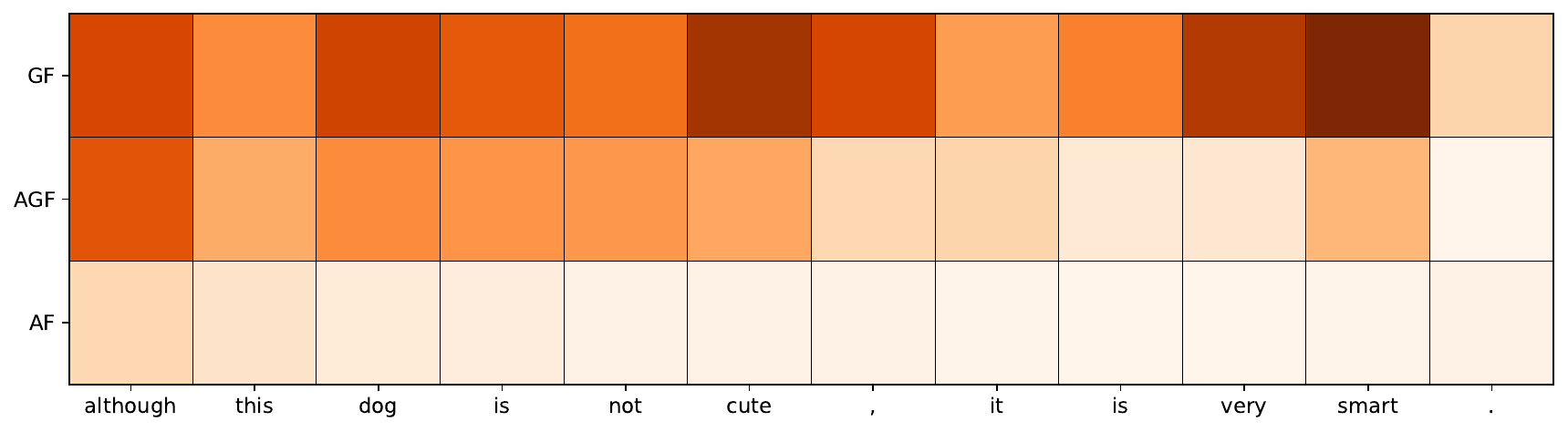}
		\label{fig5:subfig1}
	}
	\\
	\subfloat[][Normalized feature attributions for Transformer's input layer generated by \cref{algo:2} for different information tensors.]
	{
		\includegraphics[width=15cm, height=3cm]{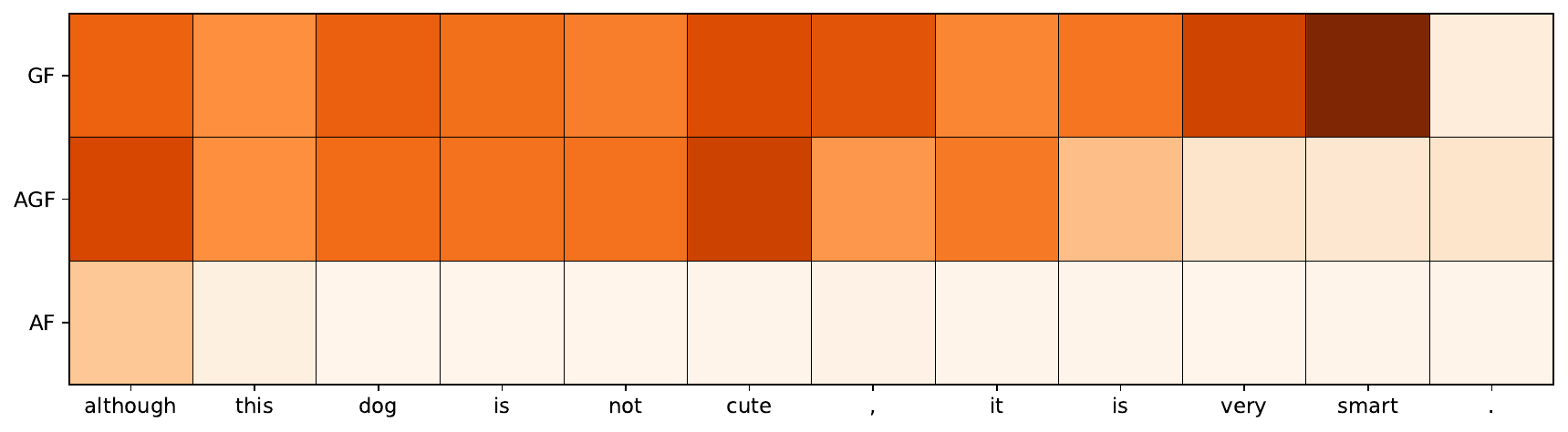}
		\label{fig5:subfig2}
	}
	\label{fig5}
	\caption{Normalized feature attributions for Transformer's input layer and different information tensors.}
\end{figure*}

\cref{fig5:subfig1} and \cref{fig5:subfig2} demo the normalized feature attributions computed for three information tensors introduced in \cref{sec:3:subsec:1} for sentiment analysis of the sentence "\textcolor{ferrari}{although this dog is not cute, it is very smart.}" using both \cref{algo:1} and \cref{algo:2}. \cref{fig5:subfig1} and \cref{fig5:subfig2} corroborate that the resulting optimal flows and their corresponding normalized attributions for the three information tensors differ depending on whether \cref{algo:1} or \cref{algo:2} is applied.

\begin{figure*}[h]
	\setcounter{subfigure}{0} 
	\centering
	\subfloat[][AF method:\cref{algo:1}]
	{
		\includegraphics[width=6.75cm, height=2.0cm]{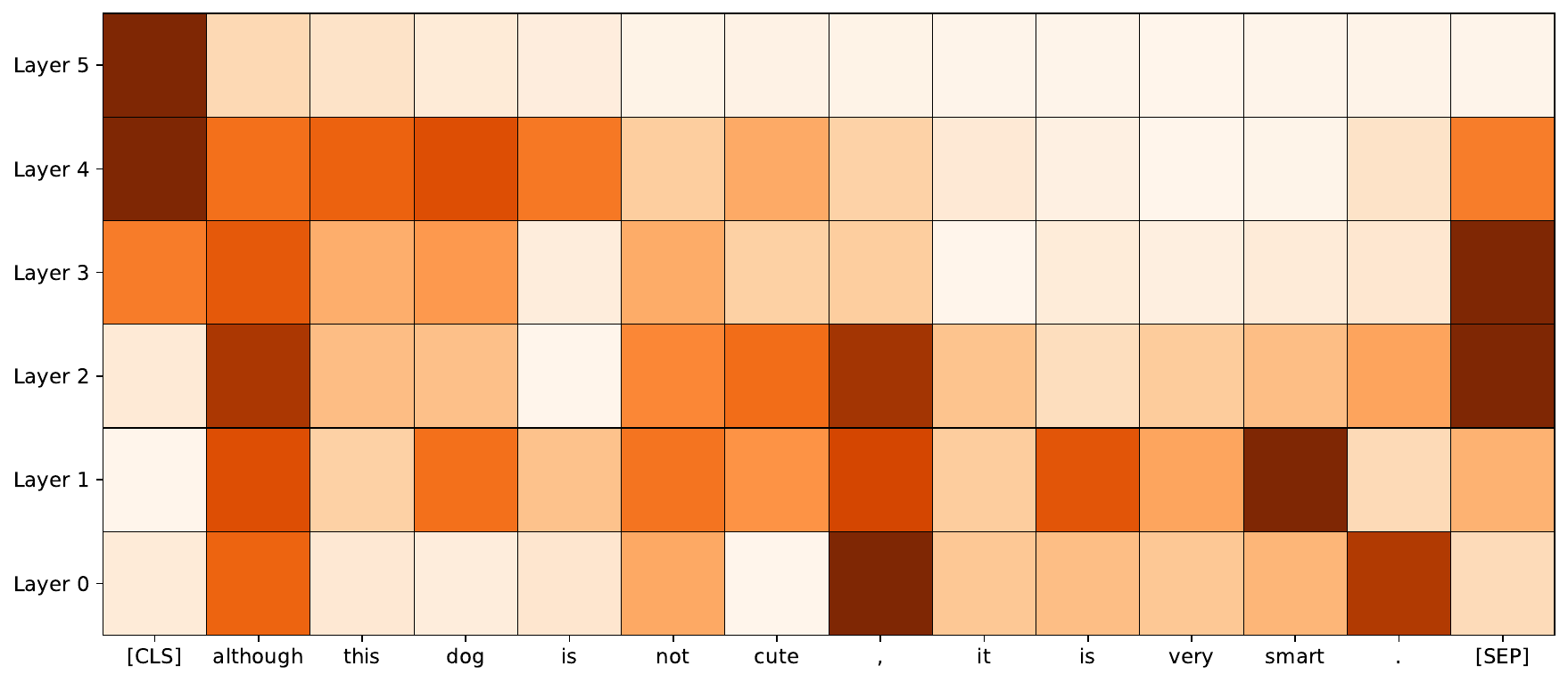}
		\label{fig6:subfig1}
	}
	\quad
	\subfloat[][AF method:\cref{algo:2}.]
	{
		\includegraphics[width=6.75cm, height=2.0cm]{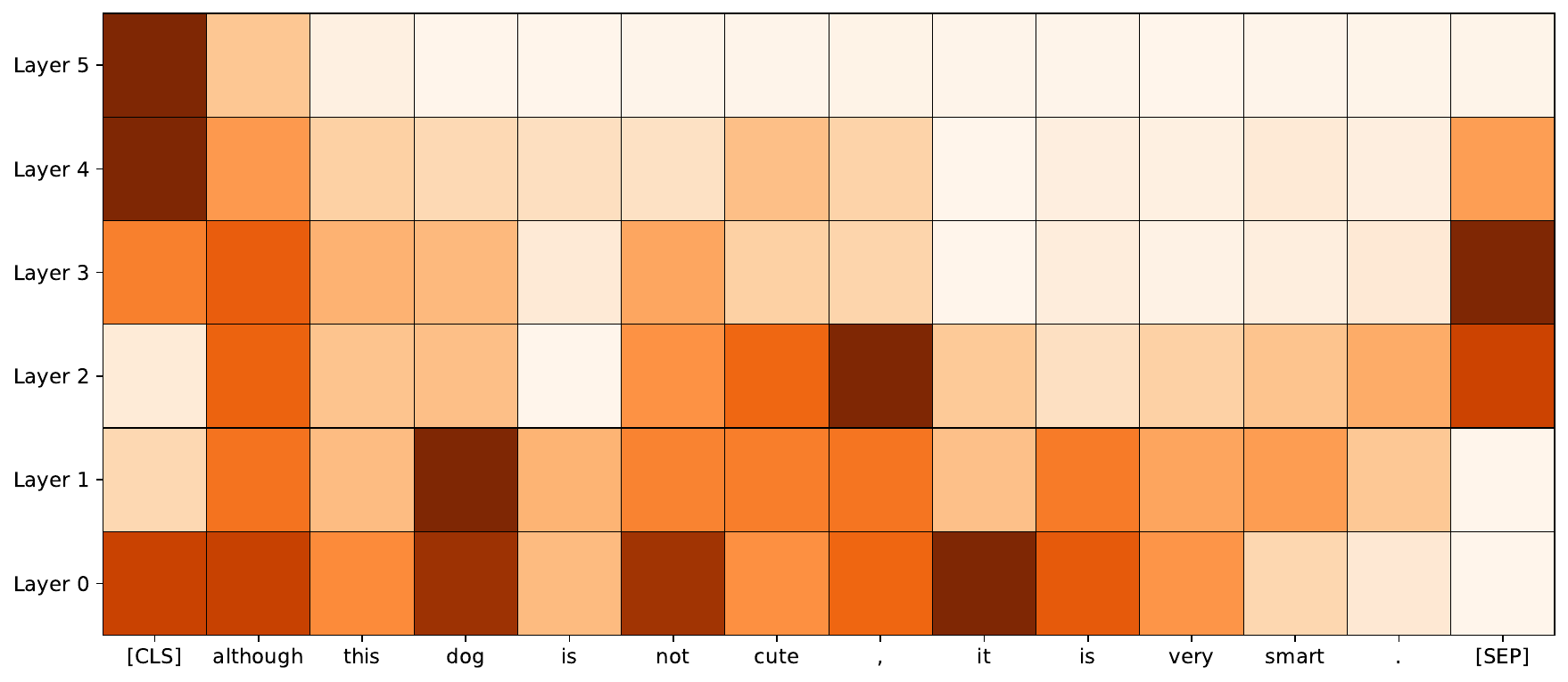}
		\label{fig6:subfig2}
	}
	\\
	\subfloat[][GF method:\cref{algo:1}.]
	{
		\includegraphics[width=6.75cm, height=2.0cm]{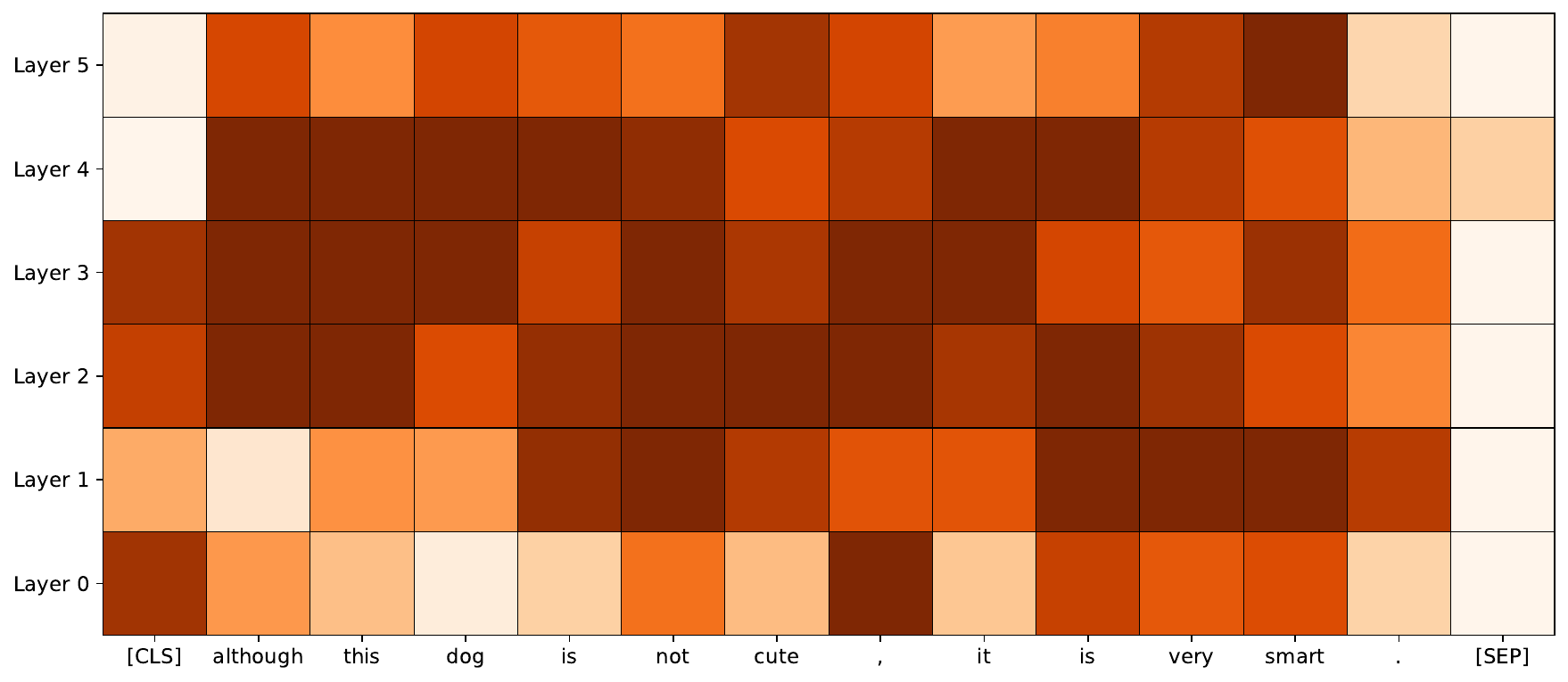}
		\label{fig6:subfig3}
	}
	\quad
	\subfloat[][GF method:\cref{algo:2}.]
	{
		\includegraphics[width=6.5cm, height=2cm]{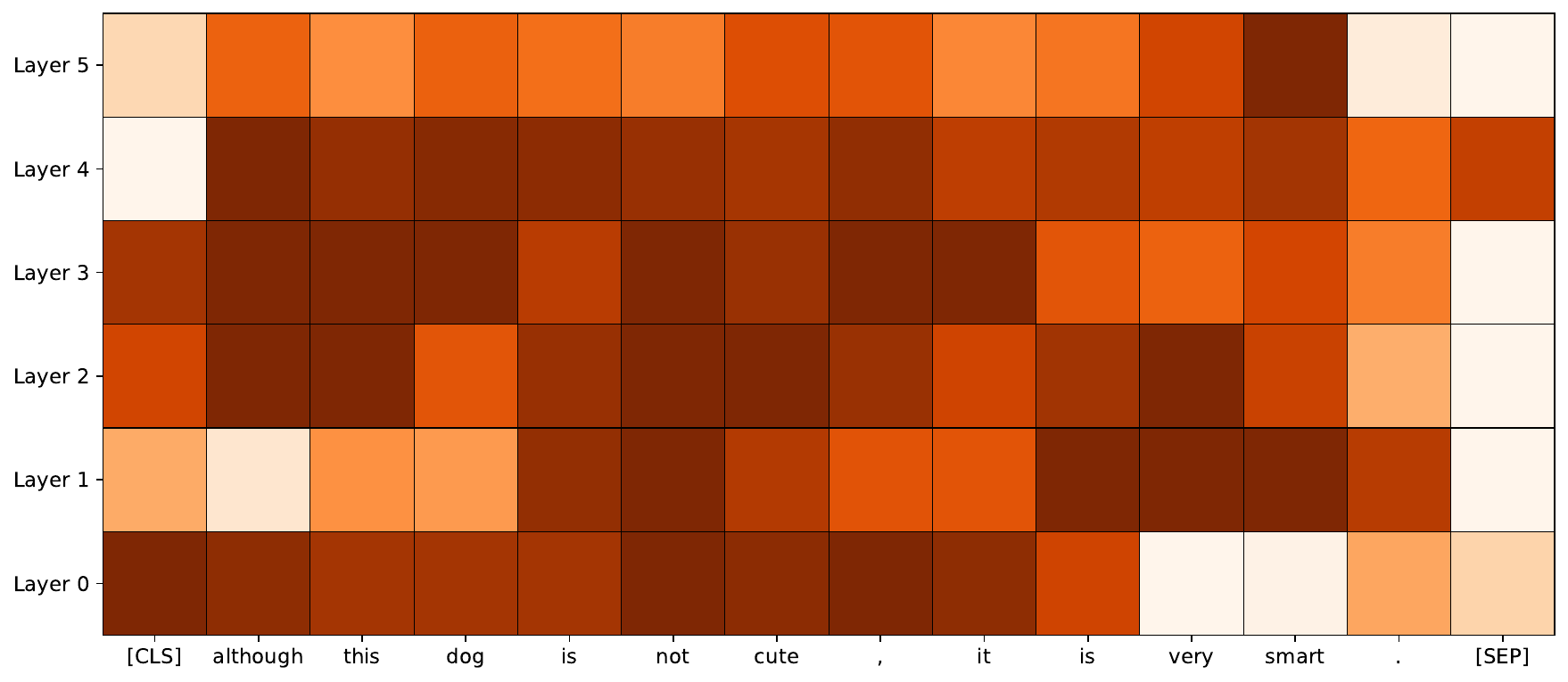}
		\label{fig6:subfig4}
	}
	\\
	\subfloat[][AGF method:\cref{algo:1}.]
	{
		\includegraphics[width=6.5cm, height=2cm]{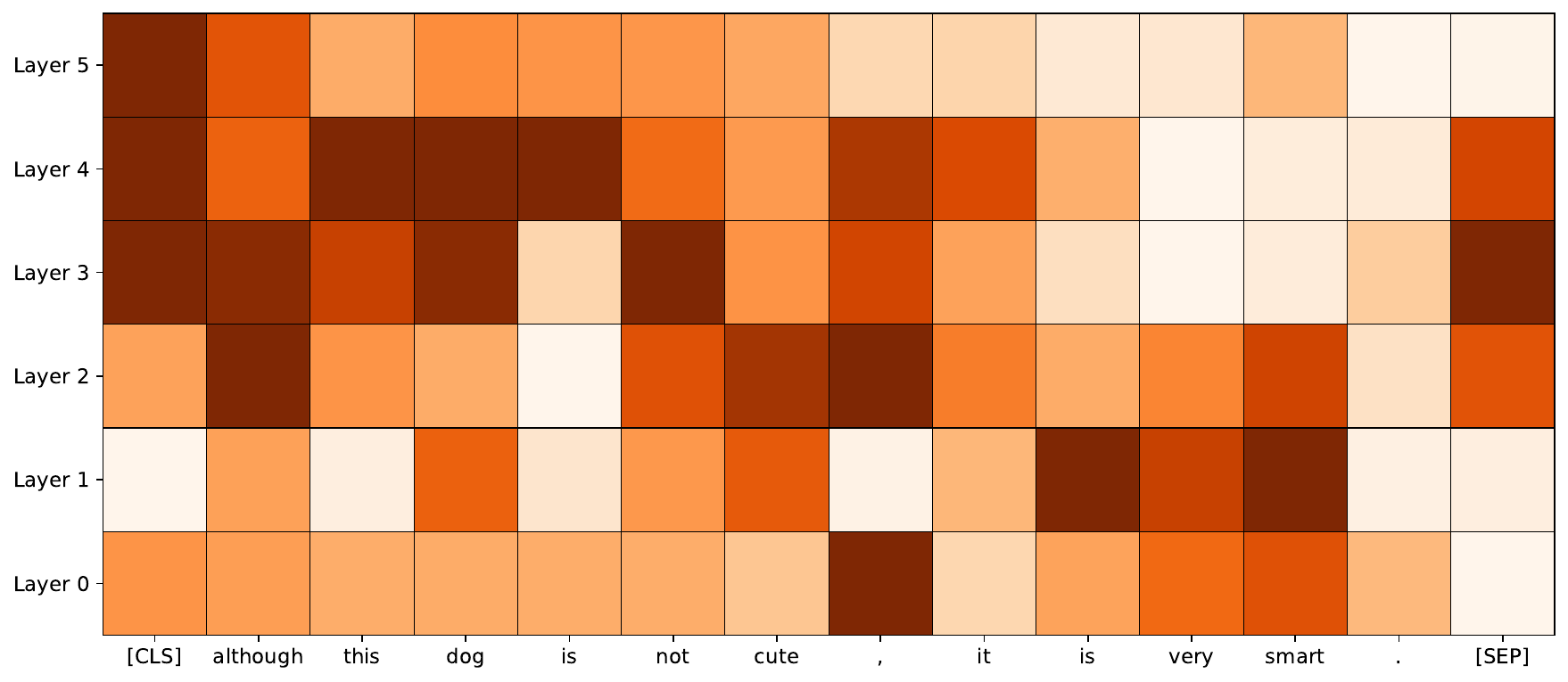}
		\label{fig6:subfig5}
	}
	\quad
	\subfloat[][AGF method:\cref{algo:2}.]
	{
		\includegraphics[width=6.5cm, height=2cm]{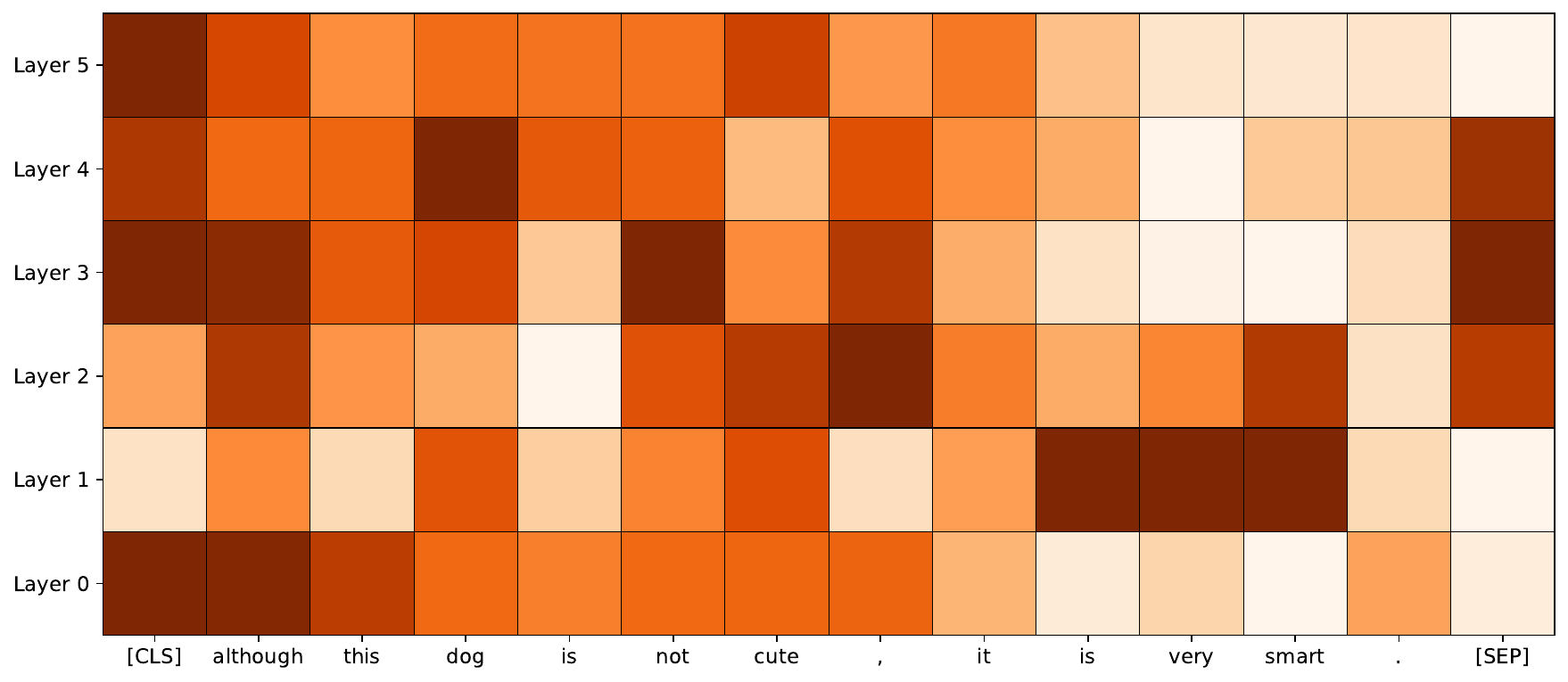}
		\label{fig6:subfig6}
	}
	\caption{Normalized feature attributions for all Transformer layers generated by \cref{algo:1} and \cref{algo:2}.}
	\label{fig6}
\end{figure*}

The layer-wise normalized feature attributions, obtained through the same process, are displayed in \cref{fig6}. For each information tensor type and layer, the resulting optimal flows and their normalized attributions differ based on whether \cref{algo:1} or \cref{algo:2} is utilized. We have also computed the optimal flow and its feature attributions for various input sentences using both algorithms for each of the information tensors AF, GF, and AGF. Our results show that the optimal flows and their corresponding feature attributions calculated using either \cref{algo:1} or \cref{algo:2} differ for all sentences.

\section{Results} \label{app:4}

\subsection{Qualitative Visualizations} \label{app:3:subapp:1}
\begin{figure*}[ht]
	\setcounter{subfigure}{0} 
	\centering
	\includegraphics[width=15.0cm, height=3.0cm]{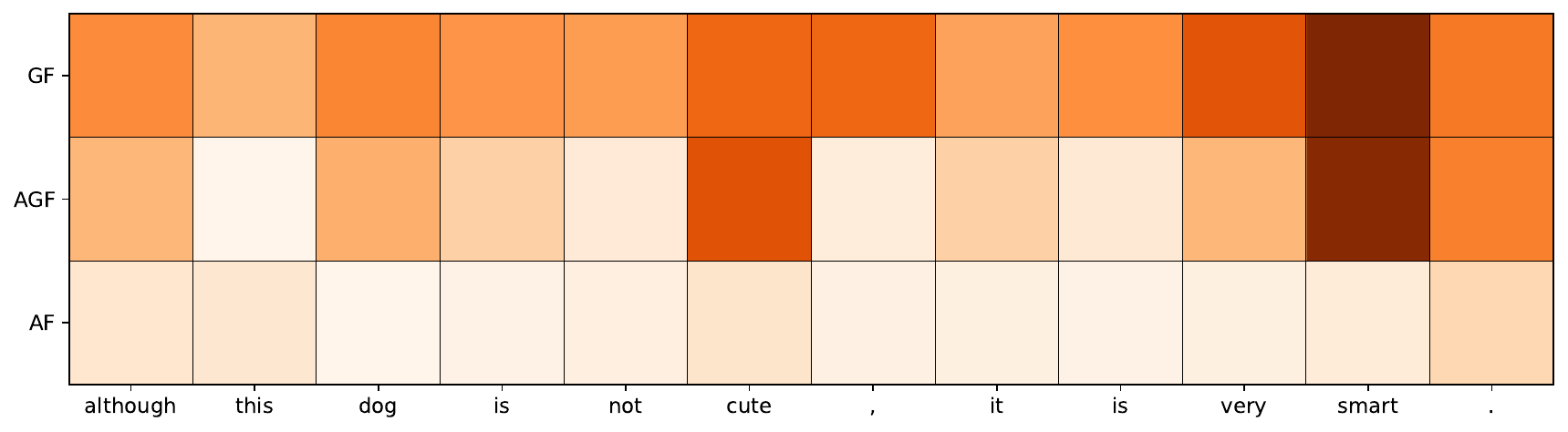}
	\caption{Visualizations of the feature attributions generated by running our proposed method on the three proposed information tensors on the showcase example.}
	\label{fig7}
\end{figure*}

This section renders a visual analysis of the feature attributions computed by our proposed methods defined in \cref{sec:3:subsec:1}. \cref{fig7} displays the feature attributions derived from our methods using graph networks generated by \cref{algo:1} or \cref{algo:2}. Notably, both approaches create identical results for both graphs. The results clearly confirm the superior performance of AGF over AF and GF, yielding more informative feature attributions. In fact, both AGF and GF accurately underscore the importance of tokens such as \textcolor{violet}{'smart'} and \textcolor{violet}{'cute'}, while assigning lower values to the less important tokens like \textcolor{deeppink}{'this'}, \textcolor{deeppink}{'it'}, and \textcolor{deeppink}{'and'}. However, AF fails to capture the expected attribution for \textcolor{violet}{'smart'} and tends to distribute attributions almost uniformly.

\subsection{Additional Results} \label{app:3:subapp:2}
\begin{figure*}[!htbp]
	\setcounter{subfigure}{0} 
	\centering
	\subfloat[][AOPC score for IMDB dataset]
	{
		\includegraphics[width=7.50cm, height=6.35cm]{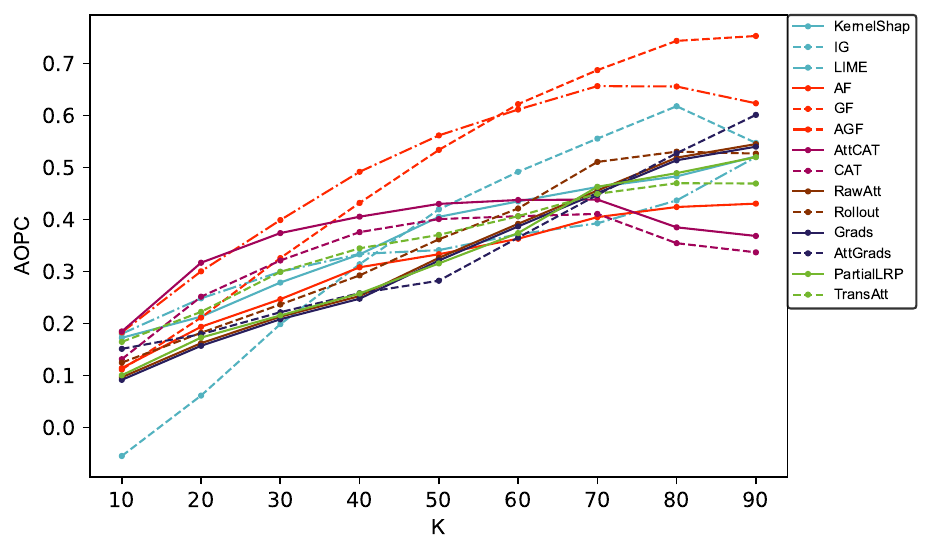}
		\label{fig8:subfig1}
	}
	\quad
	\subfloat[][LOdds score for IMDB dataset]
	{
		\includegraphics[width=7.50cm, height=6.35cm]{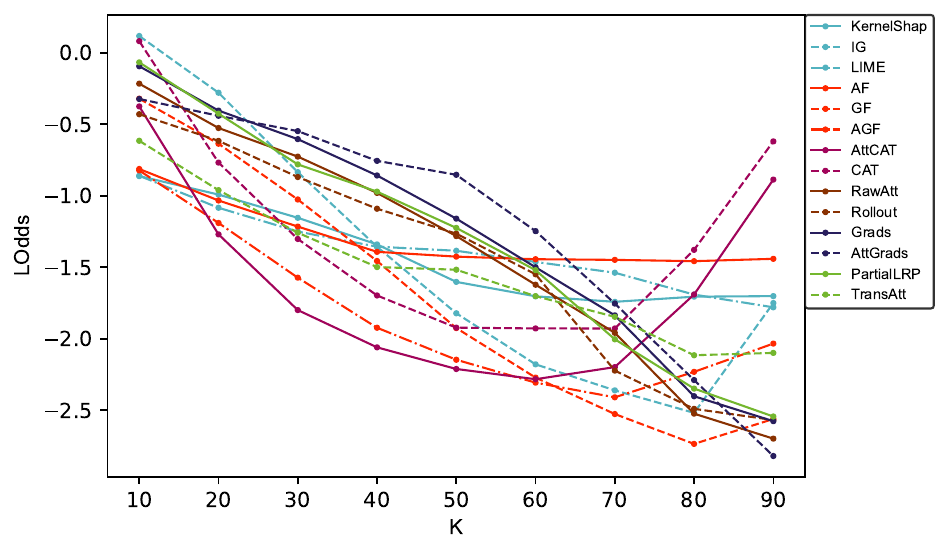}
		\label{fig8:subfig2}
	}
	\\
	\subfloat[][AOPC score for Amazon dataset]
	{
		\includegraphics[width=7.50cm, height=6.35cm]{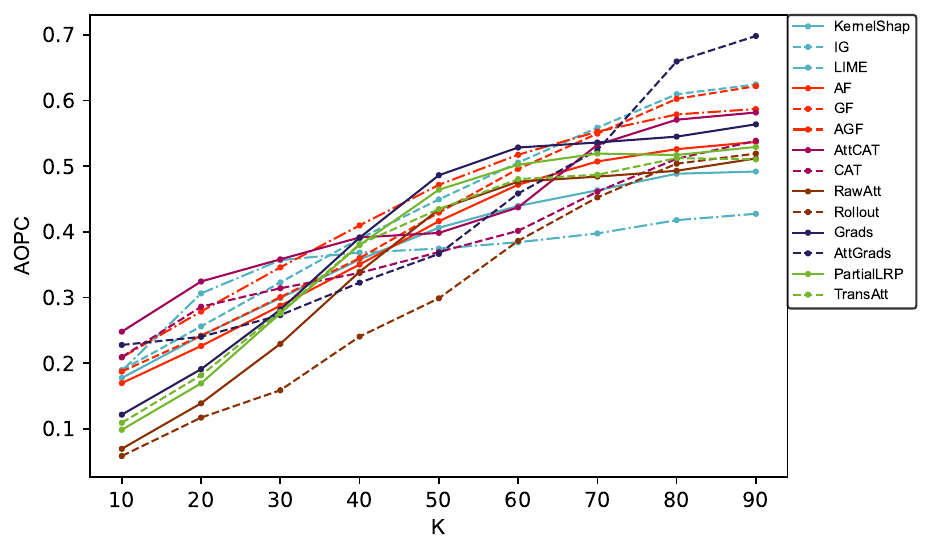}
		\label{fig8:subfig3}
	}
	\quad
	\subfloat[][LOdds score for Amazon dataset]
	{
		\includegraphics[width=7.50cm, height=6.35cm]{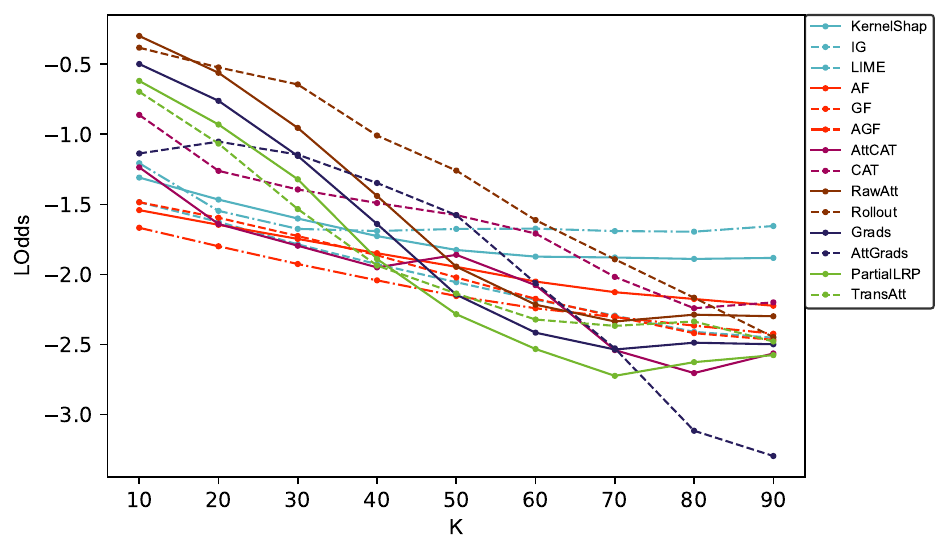}
		\label{fig8:subfig4}
	}
	\\
	\subfloat[][AOPC score for Yelp dataset]
	{
		\includegraphics[width=7.50cm, height=6.35cm]{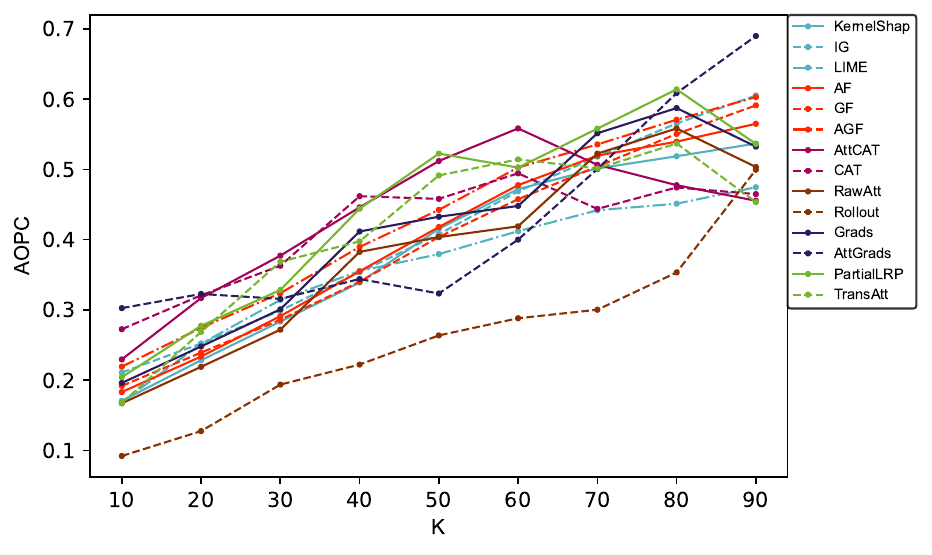}
		\label{fig8:subfig5}
	}
	\quad
	\subfloat[][LOdds score for Yelp dataset]
	{
		\includegraphics[width=7.50cm, height=6.35cm]{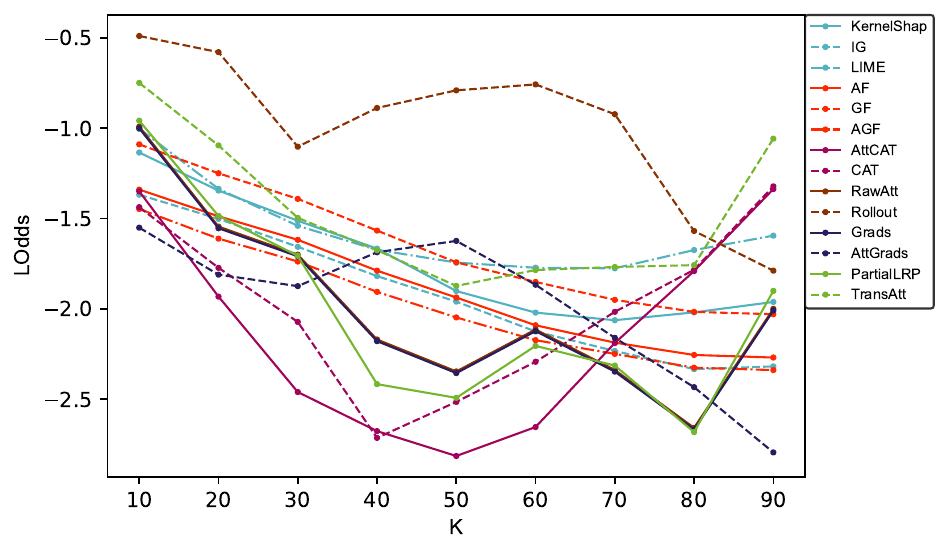}
		\label{fig8:subfig6}
	}
	
	\caption{AOPC and LOdds scores of different methods in explaining BERT across the varying corruption rates $k$ on IMDB, Amazon, and Yelp datasets. The x-axis illustrates masking the $k\%$ of the tokens in an order of decreasing saliency.}
	\label{fig8}
\end{figure*}

\cref{fig8} demonstrates a comprehensive comparison between the performance of the different feature attribution methods under varying corruption rates across three datasets: IMDB, Amazon, and Yelp. The evaluation is based on two key metrics, AOPC and LOdds, which assess how effectively each method identifies important tokens that influence model predictions. It is evident that our proposed AGF method outperform other methods by achieving the highest average AOPC and LOdds scores, notably for the IMDB and Amazon datasets. This consistently strong performance confirms AGF's robustness in identifying the most important tokens in the model's decision-making process.

\smallskip

We also evaluated various feature attribution methods using classification metrics, with results summarized in \cref{tab:3}. This table represents the average Accuracy, F1, Precision, and Recall scores across multiple $k$ values. On the SST2 dataset, AGF and GD, alongside KernelShap, achieved the highest performance. For IMDB, AGF and GF, with Integrated Gradients (IG), outperformed other methods. On Amazon, AGF and GF, combined with TransAtt, led the competition. On AG News, AGF and AF, paired with AttGrads, demonstrated superior results. The strong performance of AGF across multiple datasets and metrics underscores its versatility and reliability to identify important tokens, confirming its effectiveness as a feature attribution method in NLP models.

\smallskip

However, the Yelp dataset presents a distinct challenge where our proposed methods, including AGF, do not consistently achieve optimal results across all evaluation metrics. This performance gap is likely due to the unique characteristics of the Yelp dataset, which often contains informal language, colloquialisms, and typographical errors. The occurrence of linguistic noises in Yelp reviews is significantly higher than in other datasets like IMDB or Amazon. These textual noises introduce complexity that AGF, in its current configuration, may find challenging to address effectively.

\smallskip

\begin{figure*}[!htbp]
	\setcounter{subfigure}{0} 
	\subfloat
	{\includegraphics[width=15.0cm, height=8.5cm]{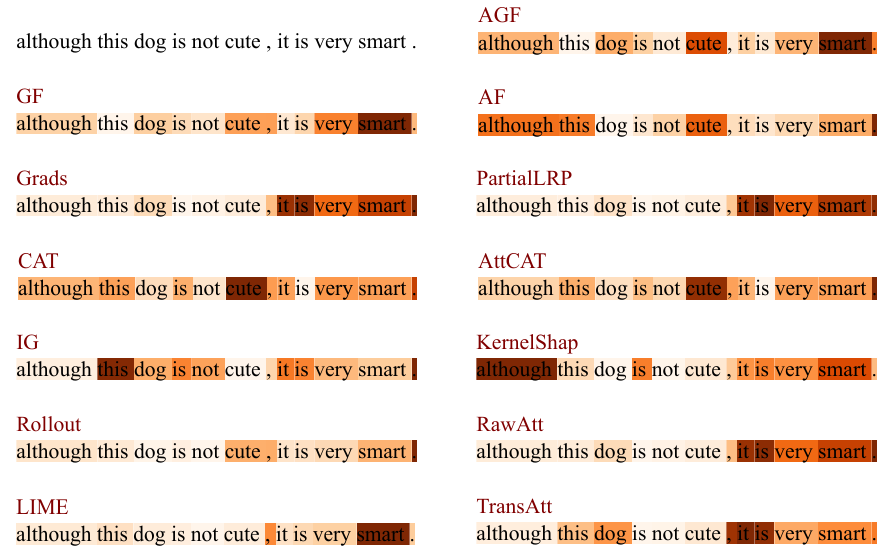}
		\label{fig9:subfig1}}
	
	\caption{Normalized feature attributions generated by the methods on the binary classification task.}
	\label{fig9}
	
\end{figure*}

\cref{fig9} compares feature attribution methods that were assessed using a model trained on the SST2 dataset, focusing on the aforementioned sentence. All methods predict a positive sentiment for the example presented. Our methods, AGF and GF, effectively identify the most important tokens, such as \textcolor{violet}{'cute'} and \textcolor{violet}{'smart'} (highlighted with dark orange shading), which play a pivotal role in the positive sentiment prediction. Some other methods, including Grads, LIME, RawAtt, and PartialLRP, also demonstrate some capability to identify important tokens. However, methods like AF, AttCAT, CAT, Rollout, KernelShap, and IG struggle to accurately identify these important tokens.

\begin{table*}[ht]
	\centering
	\scriptsize
	\setlength{\tabcolsep}{2pt}
	\begin{tabularx}{\textwidth}{|l|*{20}{>{\centering\arraybackslash}X|}}
		\toprule
		\multicolumn{1}{l}{Methods} & \multicolumn{4}{c}{SST2} & \multicolumn{4}{c}{IMDB} & \multicolumn{4}{c}{Yelp} & \multicolumn{4}{c}{Amazon} & \multicolumn{4}{c}{AG News} \\
		\cmidrule(l){2-5} \cmidrule(l){6-9} \cmidrule(l){10-13} \cmidrule(l){14-17} \cmidrule(l){18-21}
		\multicolumn{1}{l}{} & F1$\downarrow$ & Acc$\downarrow$ & Prec$\downarrow$ & Rec$\downarrow$ & F1$\downarrow$ & Acc$\downarrow$ & Prec$\downarrow$ & Rec$\downarrow$ & F1$\downarrow$ & Acc$\downarrow$ & Prec$\downarrow$ & Rec$\downarrow$ & F1$\downarrow$ & Acc$\downarrow$ & Prec$\downarrow$ & Rec$\downarrow$ & F1$\downarrow$ & Acc$\downarrow$ & Prec$\downarrow$ & Rec$\downarrow$ \\
		\hline
		\rowcolor{palegrey} RawAtt & 0.75 & 0.75 & 0.72 & 0.79 & 0.69 & 0.67 & 0.70 & 0.69 & 0.68 & 0.72 & 0.74 & 0.63 & 0.67 & 0.68 & 0.67 & 0.66 & 0.65 & 0.68 & 0.68 & 0.68 \\
		\rowcolor{palegrey} Rollout & 0.81 & 0.82 & 0.80 & 0.82 & 0.74 & 0.67 & 0.65 & 0.84 & 0.80 & 0.83 & 0.88 & 0.74 & 0.71 & 0.73 & 0.71 & 0.72 & 0.61 & 0.63 & 0.64 & 0.63 \\
		\hline
		\rowcolor{palegrey} Grads & 0.78 & 0.75 & 0.72 & 0.79 & 0.69 & 0.67 & 0.70 & 0.69 & 0.68 & 0.72 & 0.74 & 0.63 & 0.67 & 0.68 & 0.67 & 0.66 & 0.65 & 0.68 & 0.68 & 0.68 \\
		\rowcolor{palegrey} AttGrads & 0.78 & 0.78  & 0.75 & 0.82 & 0.76 & 0.75 & 0.78 & 0.76 & 0.91 & 0.91 & 0.89 & 0.93 & 0.79 & 0.82 & 0.80 & 0.78 & 0.58 & 0.61 & 0.60 & 0.60 \\
		\rowcolor{palegrey} CAT & 0.68 & 0.65 & 0.61 & 0.76 & 0.56 & 0.49 & 0.51 & 0.65 & 0.70 & 0.70 & 0.68 & 0.73 & 0.68 & 0.64 & 0.67 & 0.70 & 0.63 & 0.64 & 0.64 & 0.64 \\
		\rowcolor{palegrey} AttCAT & 0.68 & 0.65 & 0.62 & 0.76 & 0.57 & 0.48 & 0.50 & 0.64 & 0.67 & 0.66 & 0.65 & 0.70 & 0.67 & 0.62 & 0.66 & 0.68 & 0.64 & 0.64 & 0.65 & 0.64 \\
		\hline
		\rowcolor{palegrey} PartialLRP & 0.75 & 0.75 & 0.71 & 0.78 & 0.66 & 0.65 & 0.67 & 0.67 & 0.65 & 0.70 & 0.71 & 0.61 & 0.65 & 0.66 & 0.65 & 0.64 & 0.65 & 0.68 & 0.68 & 0.68 \\
		\rowcolor{palegrey} TransAtt & 0.73 & 0.72 & 0.69 & 0.76 & 0.61 & 0.58 & 0.60 & 0.62 & 0.62 & 0.66 & 0.66 & 0.60 & 0.62 & 0.63 & 0.63 & 0.61 & 0.63 & 0.66 & 0.66 & 0.66 \\
		\hline
		\rowcolor{palegrey} LIME & 0.61 & 0.63 & 0.62 & 0.63 & 0.55 & 0.55 & 0.55 & 0.55 & 0.72 & 0.73 & 0.72 & 0.73 & 0.72 & 0.73 & 0.72 & 0.73 & 0.67 & 0.67 & 0.68 & 0.67 \\
		\rowcolor{palegrey} KernelShap & 0.53 & 0.52 & 0.53 & 0.53 & 0.67 & 0.69 & 0.68 & 0.69 & 0.77 & 0.78 & 0.77 & 0.78 & 0.67 & 0.68 & 0.67 & 0.68 & 0.74 & 0.74 & 0.75 & 0.74 \\
		\rowcolor{palegrey} IG & 0.56 & 0.57 & 0.56 & 0.57 & 0.48 & 0.50 & 0.48 & 0.50 & 0.72 & 0.72 & 0.72 & 0.72 & 0.73 & 0.74 & 0.73 & 0.74 & 0.67 & 0.68 & 0.67 & 0.67 \\
		\hline
		\rowcolor{palegrey} AF & 0.72 & 0.72 & 0.71 & 0.71 & 0.65 & 0.68 & 0.70 & 0.68 & 0.71 & 0.71 & 0.71 & 0.70 & 0.69 & 0.71 & 0.71 & 0.71 & 0.64 & 0.62 & 0.65 & 0.64 \\
		\rowcolor{palegrey} GF & 0.56 & 0.57 & 0.56 & 0.56 & 0.45 & .0.48 & 0.45 & 0.48 & 0.70 & 0.71 & 0.70 & 0.71 & 0.65 & 0.67 & 0.66 & 0.67 & 0.67 & 0.68 & 0.67 & 0.67 \\
		\rowcolor{palegrey} AGF & 0.54 & 0.52 & 0.54 & 0.54 & 0.46 & 0.47 & 0.47 & 0.47 & 0.70 & 0.72 & 0.70 & 0.70 & 0.67 & 0.67 & 0.67 & 0.67  & 0.64 & 0.63 & 0.65 & 0.64 \\
		\hline
	\end{tabularx}
	\caption{The average of F1, Accuracy, Precision, and Recall scores of all methods in explaining the Transformer-based model on each dataset when we mask \textbf{top} $k\%$ tokens. Lower scores are desirable for all metrics (indicated by $\downarrow$), indicating a strong ability to mark important tokens.}
	\label{tab:3}
\end{table*}

\subsection{Statistical Significance Test} \label{app:3:subapp:3}
To perform a statistical significance test, we have employed the ASO (Almost Stochastic Order) method \citep{ulmer2022, dror2019, delbarrio2017}. This method compares the cumulative distribution functions (CDFs) of two score distributions to assess stochastic dominance. Importantly, ASO does not make any assumptions about the score distributions, which allows it to be applied to any metric where higher scores indicate better performance.

\smallskip

When comparing model $A$ with model $B$ using the ASO method, we obtain the value $\epsilon_{\min}$, which is an upper bound on the violation of stochastic order. If $\epsilon_{\min} \leq \tau$ (with $\tau \leq 0.5$), model $A$ is considered stochastically dominant over model $B$, implying superiority. This value can also be interpreted as a confidence score; a lower $\epsilon_{\min}$ signifies greater confidence in the superiority of model $A$. The null hypothesis for the ASO method is defined as follows:

\begin{equation}
	H_0: \epsilon_{\min} \geq \tau
	\label{eq:20}
\end{equation}

where the significance level $\alpha$ is an input parameter that influences $\epsilon_{\min}$. 

\smallskip

In this research, we conduct 500 independent runs for each method to perform comprehensive statistical tests, comparing the AOPC and LOdds metrics of our top-performing proposed method, AGF, against the top benchmark methods listed in \cref{tab:1} and \cref{tab:2}, using $\tau = 0.5$. As shown in \cref{tab:4} and \cref{tab:5}, the AGF method consistently outperforms these benchmark techniques across all datasets, except for the Yelp dataset.

\begin{table*}[!htbp]
	\centering
	\scriptsize
	\setlength{\tabcolsep}{4pt}

	\begin{tabularx}{0.95\textwidth}{|l|*{10}{>{\centering\arraybackslash}X|}}
		\toprule
		\multicolumn{1}{l}{Dataset}  & \multicolumn{2}{c}{SST2}   & \multicolumn{2}{c}{IMDB}  & \multicolumn{2}{c}{Yelp}  & \multicolumn{2}{c}{Amazon} &   \multicolumn{2}{c}{AG News} \\
		\cmidrule(l){2-3} \cmidrule(l){4-5} \cmidrule(l){6-7} \cmidrule(l){8-9} \cmidrule(l){10-11} 
		\multicolumn{1}{l}{}  & \multicolumn{1}{c}{AOPC}  & \multicolumn{1}{c}{LOdds}  & \multicolumn{1}{c}{AOPC}  & \multicolumn{1}{c}{LOdds}  & \multicolumn{1}{c}{AOPC}  & \multicolumn{1}{c}{LOdds} & \multicolumn{1}{c}{AOPC}  & \multicolumn{1}{c}{LOdds}  & \multicolumn{1}{c}{AOPC}  & \multicolumn{1}{c}{LOdds} \\
		\hline
		\rowcolor{palegrey} $\epsilon_{\min}$      & 0.054 & 0.039      & 0.078 & 0.061    & 0.813 & 0.924      & 0.053 & 0.043     & 0.091 & 0.076   \\
		\hline
	\end{tabularx}
	\caption{ASO test to compare AGF method with the best benchmark method when we mask \textbf{top} $k\%$ tokens.}
	\label{tab:4}
	

	\medskip
	\begin{tabularx}{0.95\textwidth}{|l|*{10}{>{\centering\arraybackslash}X|}}
		\toprule
		\multicolumn{1}{l}{Dataset}  & \multicolumn{2}{c}{SST2}   & \multicolumn{2}{c}{IMDB}  & \multicolumn{2}{c}{Yelp}  & \multicolumn{2}{c}{Amazon} &   \multicolumn{2}{c}{AG News} \\
		\cmidrule(l){2-3} \cmidrule(l){4-5} \cmidrule(l){6-7} \cmidrule(l){8-9} \cmidrule(l){10-11} 
		\multicolumn{1}{l}{}  & \multicolumn{1}{c}{AOPC}  & \multicolumn{1}{c}{LOdds}  & \multicolumn{1}{c}{AOPC}  & \multicolumn{1}{c}{LOdds}  & \multicolumn{1}{c}{AOPC}  & \multicolumn{1}{c}{LOdds} & \multicolumn{1}{c}{AOPC}  & \multicolumn{1}{c}{LOdds}  & \multicolumn{1}{c}{AOPC}  & \multicolumn{1}{c}{LOdds} \\
		\hline
		\rowcolor{palegrey} $\epsilon_{\min}$      & 0.049 & 0.037      & 0.113 & 0.085     & 0.913 & 0.824      & 0.062 & 0.053     & 0.091 & 0.067   \\
		\hline
	\end{tabularx}
	\caption{ASO test to compare AGF method with the best benchmark method when we mask \textbf{bottom} $k\%$ tokens.}
	\label{tab:5}
\end{table*}

\section{Proofs} \label{app:5}

\begin{proof}[\textbf{\cref{cor:1}}]
	While Shapley values are unique, our findings in \cref{sec:3:subsec:3} and \cref{app:3} present a critical inconsistency. We demonstrated that the optimal solution of the maximum flow problems defined in \cref{eq:8} is not necessarily unique, thereby disproving the claim that the feature attributions proposed by \citet{ethayarajh2021b} are Shapley values.
	
	\smallskip
	
	The non-uniqueness of these attributions, as evidenced by our proof, fundamentally conflicts with the defining properties of Shapley values. If these attributions defined by  \citet{ethayarajh2021b} were indeed Shapley values, they would necessarily be unique. However, our observations demonstrate that since the optimal solution of the maximum flow problem is not necessarily unique,  we can derive corresponding feature attributions from each optimal solution that differ from one another.
\end{proof}

\begin{proof}[\textbf{\cref{th:2}}]
	As the optimal flow $\bm{f}^{\ast}$ is computed once for the entire graph and not for each potential subgraph, and the players (tokens) are all disjoint without any connections in $S$, blocking the flow through one player does not impact the outflow of any other players. Therefore, for every $S \subseteq N$ where $i \! \notin \! S$, we have $|f_\text{out}(i)| = v(S \cup \{i\}) - v(S)$. Utilizing the definition of Shapley values in \cref{eq:14}, we obtain:
	\begin{equation}
		\footnotesize
		\begin{aligned}
			\phi_i(\vartheta) & =\mathlarger{\sum\limits_{S \subseteq N \backslash\{i\}}} \frac{|S| !(|N|-|S|-1) !}{|N| !}(\vartheta(S \cup\{i\})-\vartheta(S)) \\
			& =\mathlarger{\sum\limits_{S \subseteq N \backslash\{i\}}} \frac{|S| !(|N|-|S|-1) !}{|N| !}(|f_\text{out}(i)|) \\
			& = |f_\text{out}(i)|\mathlarger{\sum\limits_{S \subseteq N \backslash\{i\}}} \frac{|S| !(|N|-|S|-1) !}{|N|!} \\
			& = |f_\text{out}(i)|
		\end{aligned}
		\label{eq:21}
	\end{equation}
	It is also evident that the defined function meets all four fairness-based axioms of efficiency, symmetry, linearity, and additivity.
\end{proof}

\begin{figure}[!htbp]
	\setcounter{subfigure}{0} 
	\centering
	\includegraphics[width=8.0cm, height=5.0cm]{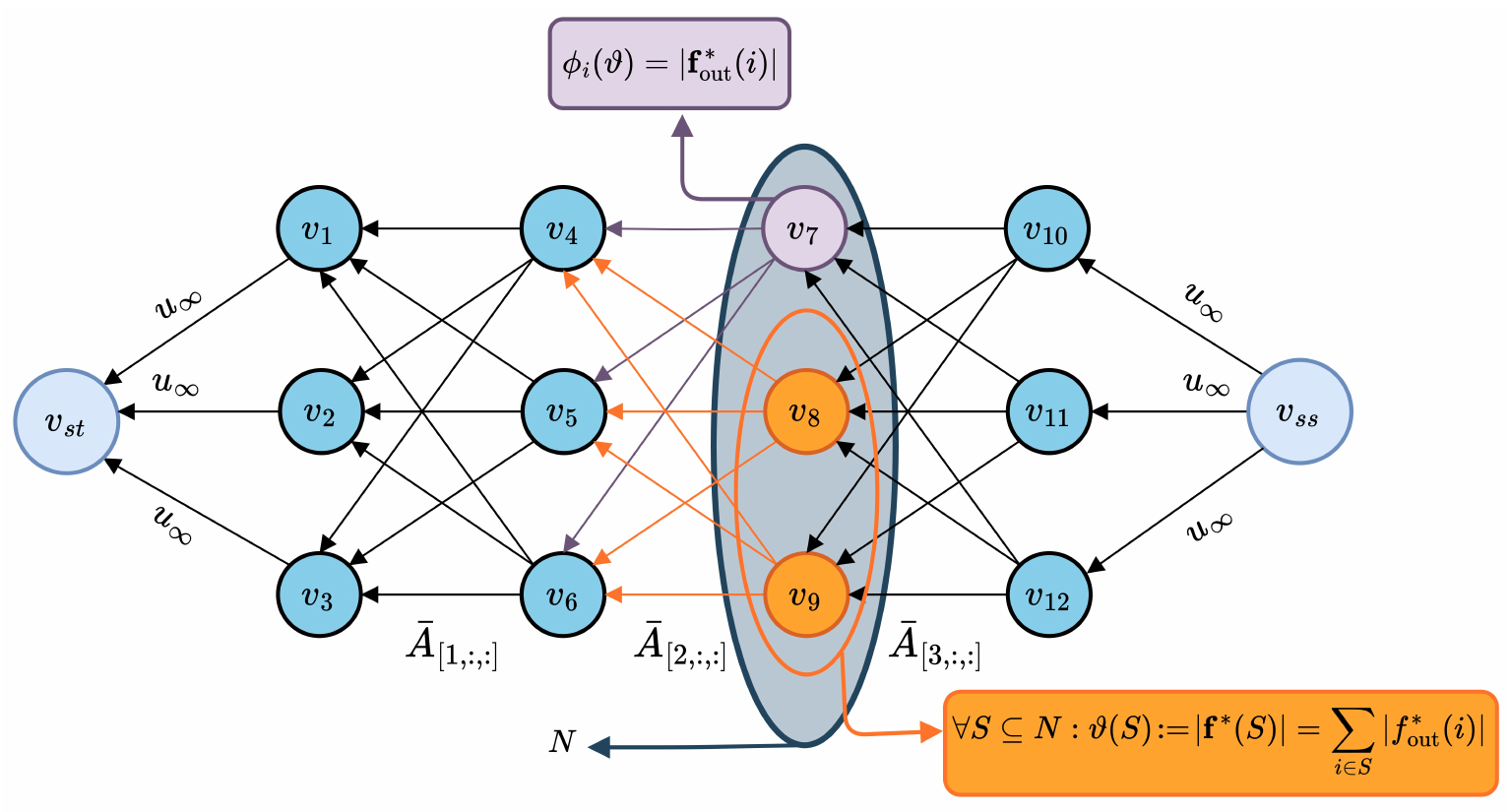}
	\caption{The procedure to define the cooperative game $(N, \vartheta)$ employing the solution of barrier-regularized maximum flow or its corresponding MCC problem.}
	\label{fig10}
\end{figure}

\section{Implementation Details} \label{app:6}

\subsection{Datasets} \label{app:6:subapp:1}

\cref{tab:6} illustrates comprehensive statistics of the datasets utilized for the classification task. We randomly selected $5,000$ sentences from each test section of the datasets, except for those with a test size less than $5,000$, where we retained all samples. Furthermore, we prioritized diversity in our sampling process by incorporating sentences of varying lengths, with an equal distribution between those shorter and longer than the mode size of the test dataset.

\begin{table*}[!htbp]
	\centering
	\scriptsize

	\begin{tabular}{|l | c | c| c |c | c| c |l|}
		\toprule
		\multicolumn{1}{l}{\textbf{Datasets}}  & \multicolumn{1}{c}{\textbf{\# Test Samples}} & \multicolumn{1}{c}{\textbf{\# Classes}} & \multicolumn{1}{c}{$\bm{\ell_{\textbf{mode}}}$} &
		\multicolumn{1}{c}{$\bm{\ell_{\textbf{min}}}$} &
		\multicolumn{1}{c}{$\bm{\ell_{\textbf{max}}}$} &
		\multicolumn{1}{c}{$\bm{\ell_{\textbf{avg}}}$} & \multicolumn{1}{l}{\textbf{Pre-trained Model}}\\
		\hline
		
		\rowcolor{palegrey}
		SST2 & 1,821 & 2 & 108 & 5 & 256 & 103.3 & \href{https://huggingface.co/textattack/bert-base-uncased-SST-2}{textattack/bert-base-uncased-SST-2} \rule[-0.5ex]{0pt}{3ex}\\
		\hline
		
		\rowcolor{palegrey}
		Amazon & 5,000 & 2 & 127  & 15 & 1009 & 404.9 & \href{https://huggingface.co/fabriceyhc/bert-base-uncased-amazon_polarity}{fabriceyhc/bert-base-uncased-amazon\_polarity} \rule[-0.5ex]{0pt}{3ex}\\
		\hline
		
		\rowcolor{palegrey}
		IMDB & 5,000 & 2 & 670 & 32 & 12988 & 1293.8 & \href{https://huggingface.co/fabriceyhc/bert-base-uncased-imdb}{fabriceyhc/bert-base-uncased-imdb} \rule[-0.5ex]{0pt}{3ex} \\
		\hline
		
		\rowcolor{palegrey}
		Yelp & 5,000 & 2  & 313 & 4 & 5107 & 723.8 & \href{https://huggingface.co/fabriceyhc/bert-base-uncased-yelp_polarity}{fabriceyhc/bert-base-uncased-yelp\_polarity} \rule[-0.5ex]{0pt}{3ex} \\
		\hline
		
		\rowcolor{palegrey}
		AG News & 5,000 & 4 & 238 & 100 & 892 & 235.3 & \href{https://huggingface.co/fabriceyhc/bert-base-uncased-ag_news}{fabriceyhc/bert-base-uncased-ag\_news} \rule[-0.5ex]{0pt}{3ex} \\
		\hline
		
	\end{tabular}
	\caption{Statistical information and the pre-trained models employed for each dataset.} \label{tab:6}
\end{table*}

\subsection{Time Complexity of Proposed Methods} \label{app:6:subapp:2}
In the minimum-cost circulation problem, we are given a directed graph $G=(V, E)$ with $|V|=n$ vertices and $|E|=m$ edges, upper and lower edge capacities $\bm{u}, \bm{l} \in \mathbb{R}^m$, and edge costs $\bm{c} \in \mathbb{R}^m$. Our objective is to find a circulation $\bm{f} \in \mathbb{R}^m$ satisfying:
\[
\underset{\substack{\bm{B}^{\top} \bm{f}=\bm{0} \\ \bm{l} \leq \bm{f} \leq \bm{u} }}{\arg \min } \hspace{2pt} \bm{c}^{\top} \bm{f}
\]

where $\mathbf{B} \in \mathbb{R}^{m \times n}$ is the edge-vertex incidence matrix. 

\smallskip

To comprehensively compare the computational efficiency between algorithms, we consider $\widetilde{\bm{l}}, \widetilde{\bm{u}}, \widetilde{\bm{c}}$ as the integral representations of the parameters $\bm{l}, \bm{u}, \bm{c}$, derived from \cref{algo:1} or \cref{algo:2} and define $U = \|\widetilde{\bm{u}}\|_{\infty}$ and $C = \|\widetilde{\bm{c}}\|_{\infty}$. \cref{tab:7} provides a comparative analysis of state-of-the-art iterative algorithms designed to solve the maximum flow and minimum-cost circulation problems. 

\smallskip

While the most efficient algorithms exhibits asymptotically near-linear runtime with respect to the number of edges $m$, its computational cost can still be significant, particularly when we have long input sequences. To solve the minimum-cost circulation problem efficiently, we have implemented the algorithm proposed by \citet{lee2014} using CVXPY \citep{agrawal2017, diamond2016}, which ensures numerical stability and computational efficiency.

\begin{table*}[!htbp]
	\centering
	\scriptsize
	\begin{tabular}{|l|c|c|c|}
		\toprule
		\multicolumn{1}{l}{\textbf{Year}}  & \multicolumn{1}{c}{\textbf{MCC Bound}} & \multicolumn{1}{c}{\textbf{Max-Flow Bound}} & \multicolumn{1}{c}{\textbf{Author}} \\
		\hline
		
		\rowcolor{palegrey} 2014     & $O\left(m \sqrt{n} \ \text{polylog}(n) \log^2(U)\right)$ & $O\left(m \sqrt{n} \ \text{polylog}(n) \log^2(U)\right)$ & \citet{lee2014} \rule[-2ex]{0pt}{6ex} \\
		\hline		
		
		\rowcolor{palegrey} 2022     & $O\left(m^{\frac{3}{2}-\frac{1}{762}}\text{polylog}(n) \log (U+C)\right)$  & $O\left(m^{\frac{10}{7}} \ \text{polylog}(n) U^{\frac{1}{7}}\right)$  & \citet{axiotis2022}	\rule[-2ex]{0pt}{6ex} \\
		\hline
		
		\rowcolor{palegrey} 2023     & $O\left(m^{\frac{3}{2}-\frac{1}{58}}\text{polylog}(n) \log^2 (U)\right)$  & $O\left(m^{\frac{3}{2}-\frac{1}{58}}\text{polylog}(n) \log^2 (U)\right)$  & \citet{vandenbrand2023} \rule[-2ex]{0pt}{6ex} \\
		\hline
		
		\rowcolor{palegrey} 2023     & $O\left(m^{1+o(1)} \log(U) \log(C) \right)$  & $O\left(m^{1+o(1)} \log(U) \log(C) \right)$  & \citet{chen2023f} \rule[-2ex]{0pt}{6ex} \\
		\hline

	\end{tabular}
			
	\caption{Overview of recent iterative algorithms for maximum flow and minimum-cost circulation problems.}
	\label{tab:7}
\end{table*}

\smallskip

This study has been conducted on a computing device running Ubuntu 20.04.4 LTS. The system is powered by Intel(R) Xeon(R) Platinum 8368 CPUs, which operate at a clock speed of 2.40 GHz. This processor features 12 physical cores and 24 threads, enabling efficient parallel computing and optimized execution of computationally intensive tasks. The graphical computations were handled by an NVIDIA RTX 3090 Ti GPU, equipped with 40 GB of dedicated VRAM, ensuring high-speed processing of deep learning and machine learning workloads. The system is also equipped with 230 GB of dedicated system memory, ensuring smooth and efficient experimentation.

\smallskip

\cref{tab:8} compares the runtime of all methods to compute feature attributions of each token in the sentence "\textcolor{ferrari}{although this dog is not cute, it is very smart.}". Methods such as RawAtt and Rollout, which depend on raw attention weights, have the shortest runtime. In contrast, other methods that require complex post-processing steps to compute feature attributions have longer runtime. As shown in \cref{tab:8}, the runtime of our proposed methods is comparable to that of other methods in this latter category.

\begin{table}[!htbp]
	\centering
	\scriptsize
	
	\begin{tabular}{|l|c|}
		\toprule
		\multicolumn{1}{l}{Methods}  & \multicolumn{1}{c}{Runtime (seconds)}  \\
		\hline
		\rowcolor{palegrey} RawAtt      & 0.123  \\
		\rowcolor{palegrey} Rollout     & 0.154 \\
		\hline
		\rowcolor{palegrey} Grads       & 1.554 \\
		\rowcolor{palegrey} AttGrads    & 1.571  \\
		\rowcolor{palegrey} CAT         & 1.684   \\
		\rowcolor{palegrey} AttCAT      & 1.660 \\
		\hline
		\rowcolor{palegrey} PartialLRP  & 1.571   \\
		\rowcolor{palegrey} TransAtt    & 1.620 \\
		\hline
		\rowcolor{palegrey} LIME    & 1.462    \\
		\rowcolor{palegrey} KernelShap  & 2.342 \\
		\rowcolor{palegrey} IG          & 2.701  \\
		\hline
		\rowcolor{palegrey} AF          & 2.301 \\
		\rowcolor{palegrey} GF          & 2.305 \\
		\rowcolor{palegrey} AGF         & 2.306 \\ \hline
	\end{tabular}
	
	\caption{Runtime of methods for the showcase example.}
	\label{tab:8}

\end{table}

\end{document}